\documentclass[lettersize,journal]{IEEEtran}
\usepackage{amsmath,amsfonts}
\usepackage{algorithmic}
\usepackage{algorithm}
\usepackage{array}
\usepackage{textcomp}
\usepackage{stfloats}
\usepackage{url}
\usepackage{verbatim}
\usepackage{graphicx}
\usepackage{cite}
\usepackage{hyperref}
\usepackage{makecell}
\usepackage{pdflscape}
\usepackage{adjustbox}
\usepackage{enumitem}
\usepackage{multirow}
\usepackage{pifont}
\usepackage{tabularx}
\usepackage{tablefootnote}
\usepackage{booktabs}

\usepackage{subcaption}

\usepackage{pgfplots}
\usepackage{xcolor}

\usepackage{makecell}

\newcommand{\etal}{\emph{et al.}}

\newcommand{\catchyname}{Type2Branch}
\newcommand{\lossyname}{Set2set}

\pgfplotsset{compat=1.18}

\newcolumntype{P}[1]{>{\centering\arraybackslash}p{#1}}

\begin{document}
\title{\catchyname: Keystroke Biometrics\\ based on a Dual-branch Architecture\\ with Attention Mechanisms and Set2set Loss}


\author{\IEEEauthorblockN{Nahuel González\IEEEauthorrefmark{1}\thanks{Email: \href{mailto:ngonzalez@lsia.fi.uba.com}{ngonzalez@lsia.fi.uba.com}}, 
Giuseppe Stragapede\IEEEauthorrefmark{2}, 
Ruben Vera-Rodriguez\IEEEauthorrefmark{2},
Ruben Tolosana\IEEEauthorrefmark{2}
}

\IEEEauthorblockA{\IEEEauthorrefmark{1}Laboratorio de Sistemas de Informacion Avanzados (LSIA), University of Buenos Aires, Argentina}

\IEEEauthorblockA{\IEEEauthorrefmark{2}Biometrics and Data Pattern Analytics (BiDA) Lab, Universidad Autonoma de Madrid, Spain}

}


\maketitle




\begin{abstract}
In 2021, the pioneering work TypeNet showed that keystroke dynamics verification could scale to hundreds of thousands of users with minimal performance degradation. Recently, the KVC-onGoing competition\footnote{\url{https://sites.google.com/view/bida-kvc/}} has provided an open and robust experimental protocol for evaluating keystroke dynamics verification systems of such scale. 
This article describes \catchyname{}, the model and techniques that achieved the lowest error rates at the KVC-onGoing, in both desktop and mobile typing scenarios. The novelty aspects of the proposed \catchyname{} include: \textit{i)} synthesized timing features emphasizing user behavior deviation from the general population, \textit{ii)} a dual-branch architecture combining recurrent and convolutional paths with various attention mechanisms, \textit{iii)} a new loss function named Set2set that captures the global structure of the embedding space, and \textit{iv)} a training curriculum of increasing difficulty. Considering five enrollment samples per subject of approximately 50 characters typed, the proposed \catchyname{} achieves state-of-the-art performance with mean per-subject Equal Error Rates (EERs) of 0.77\% and 1.03\% on evaluation sets of respectively 15,000 and 5,000 subjects for desktop and mobile scenarios. With a fixed global threshold for all subjects, the EERs are respectively 3.25\% and 3.61\% for desktop and mobile scenarios, outperforming previous approaches by a significant margin. The source code for dataset generation, model, and training process is publicly available\footnote{\url{https://github.com/lsia/tifs-type2branch}}.
\end{abstract}


\section{Introduction}
\label{sec:Introduction}

\begin{figure*}[t]
    \centering
     \includegraphics[trim={0cm 0cm 0 0cm}, width=\linewidth]{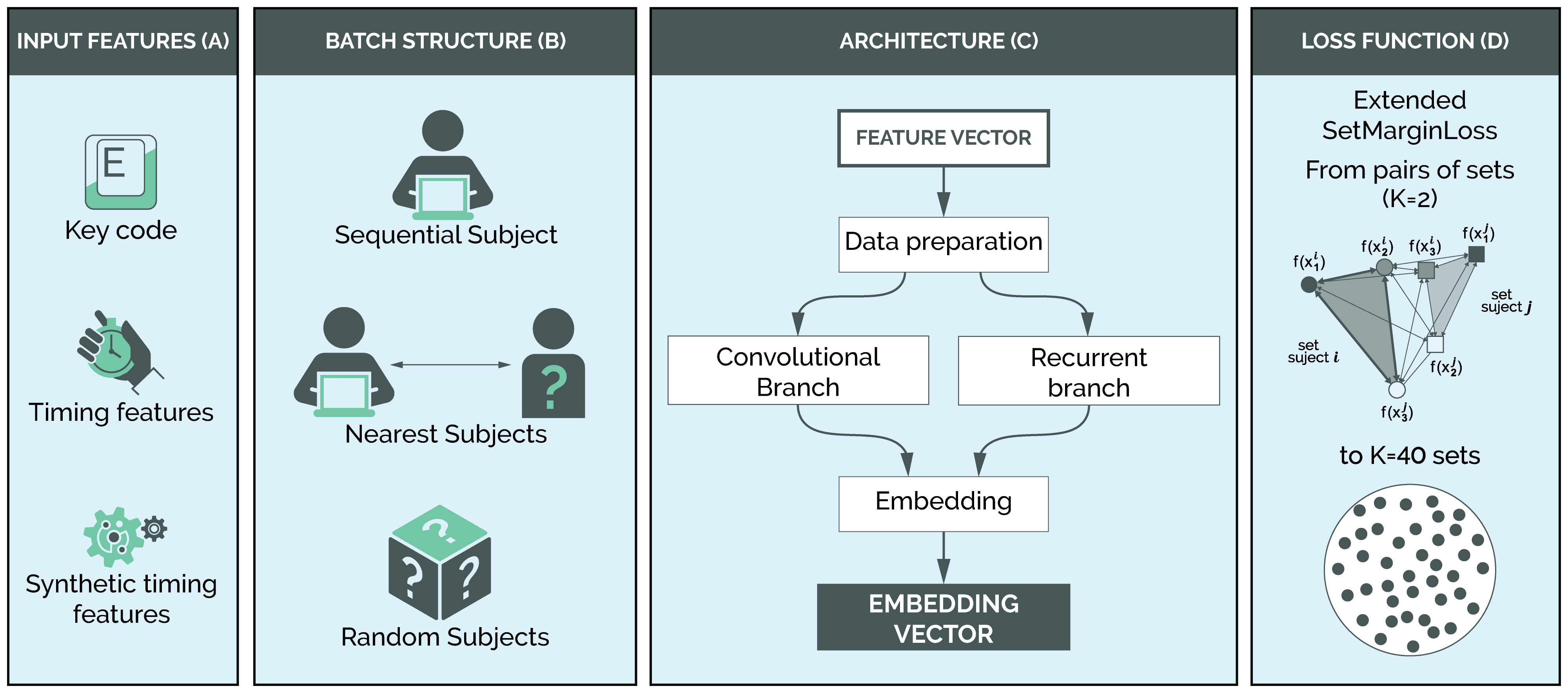}
    \caption{Overview of key aspects of \catchyname. The novel loss function, along with the proposed input features, training curriculum, and dual-branch architecture, make up the proposed biometric keystroke verification system.}
    \label{fig:workflow}
\end{figure*}

\IEEEPARstart{K}{eystroke} Dynamics (KD) refers to the typing behavior exhibited by human subjects, and it represents a form of \textit{behavioral biometric} trait, similar to gait \cite{delgado2023m}, touch gestures \cite{tolosana2020biotouchpass2, fierrez2018benchmarking}, and signature \cite{TOLOSANA2022108609}, among others. In its most basic form, keystroke dynamics is captured as discrete time events: the times at which a key is pressed and released (typically in Unix time format), along with the corresponding key code (ASCII). Additional information, such as key pressure or fingertip size, may be available based on specific hardware capabilities \cite{stragapede2023behavepassdb}. Consequently, applications based on keystroke dynamics are generally cost-effective, as they only require standard keyboards, which currently serve as the primary means for inputting textual data into digital systems, utilized by billions of users daily.

In general, behavioral biometrics, such as KD, currently does not achieve the same recognition performance as their \textit{physiological} counterparts, such as face, fingerprint, and iris. Nevertheless, it offers the advantage of operating \textit{transparently}, without requiring the subject to perform any specific procedure. Typically, biometric recognition-based security is the most prevalent application of KD. This involves both authentication scenarios: \textit{i)} verification: pairs of KD samples are captured, processed, and matched while subjects engage in activities like writing an email or taking a test on educational platforms. It can also serve as an additional layer of biometric security alongside traditional knowledge-based passwords \cite{kboc}; and \textit{ii)} identification: KD enables linking different accounts used by the same subject by matching their typing behavior among multiple samples from other subjects. Identifying or shortlisting malicious users can contribute to digital forensics applications, such as tackling toxicity, hate, and harassment on social networks \cite{mandryk2023combating}, protecting children from online grooming \cite{BORJ2023110039}, and combating fake news \cite{morales2020keystroke} and ``Wikipedia wars'' \cite{haaretz}.


The field of KD can be broadly categorized based on two criteria: \textit{i)} the type of acquisition device (keyboard), distinguishing between desktop and mobile. Mobile touchscreens tend to exhibit more variability due to differences in pose or typing activity compared to desktop keyboards; and \textit{ii)} concerning the text format, it can be classified as free, fixed, or transcript. In the case of free text, variations exist across different samples, resulting in sparser, less structured data with a higher incidence of typing errors. In contrast, fixed-text scenarios, such as an intruder typing a victims password, aim for consistent representation and lower error rates. Finally, transcript text is considered a hybrid format, involving subjects reading, memorizing, and typing a presented text. It is important to note that composition (free text) and transcription tasks produce equivalent evaluation results when used for training \cite{killourhy2012free}. As transcription is easier for the subjects, current large KD datasets like the one used in this study have almost exclusively adopted this modality for data acquisition.

In this article, we propose a new approach for KD-based biometric verification called \catchyname, featuring several novel aspects  that contribute to achieving a significant improvement in the verification performance in comparison with existing approaches in the literature. Fig. \ref{fig:workflow} provides a graphical representation describing the main novelties of the proposed \catchyname:

\begin{itemize}
    \item \textbf{Novel synthetic features (A)}. Due to the recent popularity of synthetic data to overcome challenges in biometrics \cite{tolosana2021deepwritesyn, melzi2024frcsyn}, we propose to extract synthetic timing features from the general population profile as part of the learning framework in order to allow the model to learn more subject-specific features. The synthetic features are generated with the tool reported at \cite{gonzalez2023ksdlsd}; its source code is publicly available\footnote{\url{https://github.com/SoftwareImpacts/SIMPAC-2022-276}}.
    \item \textbf{Learning curriculum (B).} We designed a learning curriculum of increasing difficulty in order to progressively show to the network the most similar subjects while including enough random sets for the model to understand the global structure of the embedding space.
    \item  \textbf{Architecture (C).} Previous research showed that keystroke timings result from a combination of two factors: a partially conscious decision process involving \textit{what} to type and an entirely unconscious motor process pertaining to \textit{how} to type \cite{gonzalez2021shape}. Based on these considerations, we propose a two-branch model architecture with self-attention modules, consisting of: \textit{i)} a convolutional branch, expected to excel at identifying common, short sequences; and \textit{ii)} a recurrent branch, expected to capture the user’s time-dependent decision process. 
    \item \textbf{Loss function (D).} Previous Distance Metric Loss (DML)-based approaches such as the SetMargin Loss proposed by Morales \textit{et al.} \cite{morales2022setmargin} extend the Triplet Loss by considering pairs of sets of samples instead of triplets. In this work, we propose the \lossyname{} Loss, which extends this idea by considering $K$ sets at a time and encouraging the network to enforce uniform class radii, leading to improved recognition performance. As detailed in Section \ref{sec:System_Description}, we use an optimized implementation of the proposed Set2set Loss for computing speed, given that a na\"ive implementation of the deeply nested loop implicit in its formulation is prohibitively slow even for small batches.
\end{itemize}

\newpage
The proposed \catchyname{} achieves the first place in both tasks of the Keystroke Verification Challenge - onGoing (KVC-onGoing)\footnote{\url{https://sites.google.com/view/bida-kvc/}} \cite{stragapede2023ieee, stragapede2023bigdata}, a public benchmark for KD-based verification in desktop and mobile scenarios (each corresponding to a task), using large-scale databases (over 185,000 subjects in total) and a standard experimental protocol. Table \ref{tab:compre_comp} provides an overview of KVC-onGoing. The proposed \catchyname{} achieves 3.33\% global Equal Error Rate (EER) in the desktop task, and 3.61\% global EER in the mobile task. Considering subject-specific comparison decision thresholds (see details in Sec. \ref{subsec:evaluation_description}), the EERs achieved by \catchyname{} are further reduced to 0.77\% and 1.03\%, respectively.  

\begin{table}[b!]
\caption{\small Comparison of state-of-the-art KD-based biometric verification systems in terms of global EER (\%) on the KVC-onGoing benchmark. \textit{D} stands for desktop, \textit{M} stands for mobile.}
\centering
\renewcommand{\arraystretch}{2}
\begin{tabular}
{c|c|c|c|c}
\makecell[c]{\textbf{System}} & \makecell[c]{\textbf{Architecture}} & \makecell[c]{\textbf{Loss}} & \makecell[c]{\textbf{\textit{D}}} & \makecell[c]{\textbf{\textit{M}}}\\
\hline
\makecell{TypeFormer\\ \cite{typeformer}} & \makecell{Modified\\Transformer} & Triplet & 12.75 & 9.45 \\
BioSense & \makecell{CNN with\\ Att. Mechanism} & \makecell{Cross-\\entropy} & 10.85 & 11.83 \\
Challenger & Transformer & Triplet & 6.79 & 5.19 \\
\makecell{TypeNet\\ \cite{acien2021typenet}} & \makecell{RNN} & Triplet & 6.76 & 13.95 \\
YYama & \makecell{Transformer\\+CNN} & \makecell{Contrastive+\\Cross-entropy} & 6.41 & 4.16 \\
\makecell{U-\\CRISPER} & \makecell{GRU-Based\\Siamese} & Triplet & 6.19 & 8.76 \\
\makecell{Keystroke\\Wizards} & \makecell{GRU (\textit{D}),\\Transformer (\textit{M})} & Triplet & 5.22 & 5.83 \\
VeriKVC & CNN & ArcFace & 4.03 & 3.78 \\
\makecell{\textbf{\catchyname{}}\\ \textbf{(LSIA)}} & \makecell{\textbf{CNN+RNN with}\\ \textbf{Att. Mechanism}} & \textbf{\lossyname{}} & \textbf{3.33} & \textbf{3.61} \\
\end{tabular}
\label{tab:compre_comp}
\end{table}

A high-level description of \catchyname{} was provided in the KVC summary paper \cite{stragapede2023bigdata}, alongside the description of the other participating teams' systems. 
The present article provides an in-depth description of the novel aspects introduced and a more extensive experimental evaluation, providing deeper insights not included in the competition summary paper. Specifically, Sec. \ref{sec:Related_Work} provides a revision of related work in the literature. Sec. \ref{sec:System_Description} describes our proposed \catchyname, including: \textit{i)} the generation of synthetic timing features used by the model to learn how to distinguish each subject from the general population profile (Sec. \ref{subsec:inputFeatures}); \textit{ii)} the batch structure and the learning curriculum of increasing difficulty (Sec. \ref{subsec:batchStructure}); \textit{iii)} the novel two-branch model architecture proposed (Sec. \ref{subsec:modelArchitecture}); \textit{iv)} the motivation (Sec. \ref{subsec:lossFunctionMotivation}) and formulation (Sec. \ref{subsec:lossFunctionFormulation}) of the \lossyname{} loss function. Then, after describing in Sec. \ref{sec:Experimental_Protocol} the experimental protocol adopted, we provide in Sec. \ref{sec:Experimental_Results} a thorough experimental evaluation of the designed system using the KVC and popular external databases. Moreover, an ablation study is carried out in Sec. \ref{subsec:ablationStudy}, shedding light on the impact of the different components of \catchyname. Also, different training components are evaluated in Sec. \ref{subsec:otherTrainingComponents}, to assess: the impact of synthetic features and learning curriculum; a comparison of the \lossyname{} Loss function with the Triplet Loss and the SetMargin Loss; the impact of varying the number of sets considered at a time $K$ by the \lossyname{} Loss function, and its $\beta$ hyperparameter, responsible for weighting the loss function radius penalty term; and the impact of increasing the number of training subjects. Finally, the limitations of the current work are discussed in Sec. \ref{subsec:limitations}.

\section{Related Work}
\label{sec:Related_Work}


In comparison with traditional handcrafted algorithms developed in the early days of KD, when computing resources were not as powerful as today, the introduction of deep learning has led to significant performance improvements in KD biometric recognition \cite{acien2021typenet}. In this section, we focus on deep learning approaches, which are more relevant to our work. For a complete literature review about KD, we invite the reader to consult \cite{maiorana2021mobile, roy2022systematic}.

One of the first studies that demonstrated that a deep neural network could improve the performance of handcrafted algorithms was based on the CMU database \cite{Deng2013, cmu}. Approaches based on neural networks were also used for auxiliary tasks aimed at enhancing authentication performance, such as predicting missing digraphs by analyzing the relationships between keystrokes \cite{Ahmed2013}. A Convolutional Neural Network (CNN) coupled with Gaussian data augmentation for the fixed-text scenario was introduced in \cite{ceker2017}, while a neural network was applied to RGB histograms derived from fixed-text keystroke in \cite{ALPAR2014213}. Multi-Layer Perceptron (MLP) architectures have also been investigated \cite{stylios2022}. In \cite{lu2019}, a concatenation of convolutional and recurrent neural networks (RNNs) was designed to extract higher-level keystroke features from the SUNY Buffalo database \cite{sun2016shared}. 
Different types of RNNs are widely used in keystroke biometrics, as seen in \cite{math10162912} (bidirectional RNN), or in \cite{Li2022}, where keystroke sequences are structured as image-like matrices and processed by a CNN combined with a Gated Recurrent Unit (GRU) network.

Generally speaking, the proliferation of machine learning algorithms capable of analyzing and learning human behaviors thrives on large-scale databases. To this end, the Aalto databases, proposed in two popular Human-Computer Interaction (HCI) studies on people's typing behavior on desktop \cite{Dhakal2018} and mobile devices \cite{palin2019people}, are extremely useful. These databases were collected by the User Interfaces\footnote{\href{https://userinterfaces.aalto.fi/}{https://userinterfaces.aalto.fi/}} group of Aalto University (Finland). The desktop\footnote{\href{https://userinterfaces.aalto.fi/136Mkeystrokes/}{https://userinterfaces.aalto.fi/136Mkeystrokes/}} \cite{Dhakal2018} database comprises around 168,000 subjects, while the mobile\footnote{\href{https://userinterfaces.aalto.fi/typing37k/}{https://userinterfaces.aalto.fi/typing37k/}} \cite{palin2019people} one encompasses approximately 37,000 subjects. As can be seen in Table \ref{tab:old_dbs}, the size of such databases is significantly greater than other public databases, effectively reflecting the challenges associated with current massive application usage.

\begin{table}[t]
\centering
\caption{\small Some of the most important public KD databases in chronological order.}
 \begin{tabular}{c|c|c|c|c} 
 \textbf{Database} & \textbf{Scenario}
 & \makecell{\textbf{No. of}\\\textbf{Subjects}} & \makecell{\textbf{Text}\\\textbf{Format}} & \makecell{\textbf{Strokes per}\\\textbf{Subject}} \\ 
 \hline
 \makecell{GREYC\\ (2009) \cite{greyckeytroke}}  & Desktop & 133 & Fixed & $\sim$800 \\
\makecell{CMU\\ (2009) \cite{cmu}} & Desktop & 51 & Fixed & $\sim$400\\ 
\makecell{BiosecurID\\ (2010) \cite{biosecurid}} & Desktop & 400 & Free & $\sim$200 \\
\makecell{KM\\ (2012) \cite{killourhy2012free}} & Desktop & 20 & \makecell{Transcript,\\ free} & ~8250 \\
\makecell{RHU\\ (2014) \cite{rhu}} & Desktop & 53 & Fixed & $\sim$600 \\ 
\makecell{Clarkson I\\ (2014) \cite{clarksonI}} & Desktop & 39 & \makecell{Fixed, free} & $\sim$20k\\
\makecell{PROSODY\\ (2014) \cite{banerjee2014keystroke}} & Desktop & 400 & \makecell{Transcript,\\ free} & ~10k \\
\makecell{SUNY\\ (2016) \cite{sun2016shared}} & Desktop & 157 & \makecell{Transcript,\\ free} & \makecell{$\sim$17k} \\ 
\makecell{Clarkson II\\ (2017) \cite{clarksonII}} & Desktop & 103 & Free & $\sim$125k \\
 \makecell{\textbf{Aalto Desktop}\\ \textbf{(2018) \cite{Dhakal2018}}} & \textbf{Desktop} & \textbf{168k} & \textbf{Transcript} & \makecell{$\sim$\textbf{750}}\\
 \makecell{\textbf{Aalto Mobile}\\ \textbf{(2019) \cite{palin2019people}}} & \textbf{Mobile} & \textbf{37k} & \textbf{Transcript} & \makecell{$\sim$\textbf{750}}\\
 \makecell{HuMIdb\\ (2020) \cite{acien2021becaptcha}} & Mobile & 600 & Fixed & \makecell{$\sim$20} \\
 \makecell{BehavePassDB\\ (2022) \cite{stragapede2023behavepassdb}} & Mobile & 81 & Free & \makecell{$\sim$100} \\
\makecell{LSIA\\ (2023) \cite{gonzalez2023datasetLSIA}} & Desktop & 135 & \makecell{Free,\\ synthesized} & ~55k \\
 \end{tabular}
 \label{tab:old_dbs}
\end{table}

The approach TypeNet proposed by Acien \textit{et al.} in \cite{acien2021typenet} is one of the first studies that adopted Aalto databases \cite{Dhakal2018} for the purpose of biometric recognition. In addition to this, a novel aspect of their work is the introduction of Long Short-Term Memory (LSTM) RNNs trained with Triplet Loss \cite{schultz2003learning}, according to an open-set learning protocol, meaning the subjects in the development and final evaluation sets are distinct. 
 In addition, they analyzed to what extent deep learning models are able to scale in keystroke biometrics to recognize subjects from a large pool while attempting to minimize the amount of data per subject required for enrollment. TypeNet was able to verify subjects' identities when the amount of data per subject is scarce, i.e., only 5 enrollment samples and 1 test sample, with 50 characters typed per sample. This is possible as TypeNet maps input data into a learned representation space that reveals a ``semantic'' structure based on distances. Such approach is known as Distance Metric Learning (DML). In \cite{morales2022setmargin}, a novel DML method, called SetMargin Loss (SM-L), was proposed to address the challenges associated with transcript-text keystroke biometrics in which the classes used in learning and inference are disjoint. SM-L is based on a learning process guided by pairs of sets instead of pairs of samples, as contrastive or Triplet Loss consider, allowing to enlarge inter-class distances while maintaining the intra-class structure of keystroke dynamics. This led to improved recognition perfomance with TypeNet.

Later on, replicating the same experimental protocol as \cite{acien2021typenet}, Stragapede \textit{et al.} proposed TypeFormer \cite{typeformer}. TypeFormer was developed starting from the Transformer model \cite{vaswani2017attention}, with several adaptations to optimize its recognition performance for KD. The model consists of  Temporal and Channel Modules enclosing two LSTM RNN layers, Gaussian Range Encoding (GRE), a multi-head Self-Attention mechanism, and a Block-Recurrent structure. In several experiments, TypeFormer outperformed TypeNet in the mobile environment, but not in the desktop case \cite{stragapede2023ieee}.

Recently, in \cite{neacsu2023doublestrokenet}, a novel approach called DoubleStrokeNet was proposed, which recognizes subjects using bigram embeddings and a Transformer that distinguishes between different bigrams. Additionally, self-supervised learning techniques are used to compute embeddings for both bigrams and subjects. The authors experimented with the Aalto databases, reaching very competitive results in terms of recognition performance. In such cases, while the ideas presented are very interesting, it is often difficult to compare the results across different studies, as different experimental settings are adopted. To this end, the first attempt to promote reproducible research and establish a baseline in biometric recognition using KD was carried out in 2016 in the form of a competition for KD by Morales \textit{et al.} \cite{kboc}, namely Keystroke Biometrics Ongoing Competition (KBOC). 
In that case, the dataset used consisted of keystroke sequences (fixed text) from 300 subjects acquired in 4 different sessions.
Following this line of research, the Keystroke Verification Challenge - onGoing (KVC-onGoing)\footnote{\href{https://sites.google.com/view/bida-kvc/}{https://sites.google.com/view/bida-kvc/}} was recently launched, considering both Aalto databases (the desktop database was used for Task 1, the mobile one for Task 2). 
The ongoing challenge is hosted on CodaLab\footnote{\href{https://codalab.lisn.upsaclay.fr/competitions/14063}{https://codalab.lisn.upsaclay.fr/competitions/14063}}. In addition, thanks to the demographic (age, gender) labels present in the original database, an analysis of the demographic variability in the scores was also carried out for purposes such as privacy quantification and fairness, alongside a thorough evaluation of the biometric verification performance. TypeNet and TypeFormer were also benchmarked on the KVC-onGoing. 
Some of the participating teams were able to outperform both TypeNet and TypeFormer on both scenarios. In particular, the approach presented in this article, \catchyname, holds the first position in both desktop and mobile tasks in the KVC-onGoing. For more information about the details of the competition, we invite the reader to consult \cite{stragapede2023ieee, stragapede2023bigdata}.

\subsection{Generation of Synthetic Features}
\label{subsec:gen_synthetic_features}

Biometric verification systems based on KD have traditionally been assessed under a zero-effort attack model. In other words, biometric samples captured from different subjects are compared, but no effort is made to emulate the characteristics of genuine subjects. To this end, recent studies have demonstrated that attacks employing statistical models and synthetic forgeries can yield significant success rates \cite{stefan2012robustness, 7358795}, raising concerns that zero-effort approaches are overly optimistic.  

In \cite{gonzalez2022towards}, the authors explored spoofing techniques leveraging higher-order contexts and empirical distributions to generate artificial samples of keystroke timings to improve existing attacks. The synthetic samples for the datasets \cite{killourhy2012free} and \cite{banerjee2014keystroke} are available at \cite{gonzalez2023datasetLSIA}.



In our proposed \catchyname,  synthetic data is incorporated as part of the input features to represent the average typing behavior of the entire training population, with the objective of reflecting how the behavior of each subject shows distinguishable patterns in comparison with the population profile. We adopt the implementation presented in \cite{gonzalez2023ksdlsd}.

\section{\catchyname: Proposed System}
\label{sec:System_Description}
%
%
Fig. \ref{fig:workflow} shows an overview of \catchyname{} and its training process. The source code is publicly available in GitHub\footnote{\url{https://github.com/lsia/tifs-type2branch}}. We describe next the key modules of our proposed keystroke verification system.

%
%

\begin{table}[]
    \centering
    \caption{Terminology and definitions considered in the present article.}
    \begin{tabular}{|c|c|}
        \hline
        \textbf{Notation}  & \textbf{Description}\\   
        \hline
        \hline
        \multicolumn{2}{|c|}{\textbf{Keystroke Features}} \\
        \hline
        $k_i$ & Key code \\
        $t_i^P$ & Key press timestamp \\
        $t_i^R$ & Key release timestamp \\
        $t_i^{HT}$ & Hold time (press/release) \\
        $t_i^{FT}$ & Flight time (press/press) \\        
        $s_i^{HT}$ & Synthetic hold time (press/release) \\
        $s_i^{FT}$ & Synthetic flight time (press/press) \\        
        \hline
        \hline        
        \multicolumn{2}{|c|}{\textbf{Batch Composition}} \\
        \hline
        $N$ & Samples per set/user \\        
        $K$ & Sets/users per batch \\        
        $w_n^k$ & $n$--th sample of the $k$--th user \\        
        \hline
        \hline        
        \multicolumn{2}{|c|}{\textbf{Embeddings}} \\
        \hline
        $x_n^k$ & Embedding of $w_n^k$ \\     
        $\mu(k)$ & Center of the $k$--th user embeddings \\        
        $r(k)$ & Mean radius of the $k$--th user embeddings \\   
        $R$ & Mean radius over all classes \\  
        \hline
    \end{tabular}
    \label{tab:notation}
\end{table}

\subsection{Terminology and Definitions}

For reference, a summary of the terminology and definitions considered in the article is provided in Table \ref{tab:notation}. A KD sample $\mathbf{w}$ is a sequence $\mathbf{w}_1, \ldots, \mathbf{w}_M$, of fixed length $M$, of tuples of the form
$$(k_i, \, t^{P}_i, \, t^{R}_i)$$
where $k_i$ is the integer key code corresponding to the $i$--th keystroke, $t^{P}_i$ is the timestamp of its key press event, and $t^{R}_i$ is the timestamp of its key release event.

A class is given by the set of samples of a single subject. We assume that the number $N$ of samples in each class, is fixed. A batch consists of $NK$ samples, where all the $N$ samples of $K$ selected classes are included. The $n$--th sample of the $k$--th class is denoted as $\mathbf{w}_n^k$ and its corresponding embedding vector as $\mathbf{x}_n^k$. The center of the embeddings for the $k$--th set is denoted as $\mathbf{\mu}(k)$. The mean radius of the $k$--th set embeddings cluster is denoted by $r(k)$. Using the former, the \emph{radius penalty} for a batch is defined as
\begin{align}
\label{eq:radiuspenalty}
\mathcal{L}_{RP} = \frac{1}{K} \sum_{k=1}^K \left| \frac{r(k)}{R} - 1 \right|
\end{align}
where the term $R$, defined as
\begin{align}
\text{$R$} = \frac{1}{K} \sum_{k=1}^K r(k)
\end{align}
is the mean radius over all classes.

\begin{figure*}[t]
    \centering
     \includegraphics[trim={0cm 0cm 0 0cm}, width=0.7\linewidth]{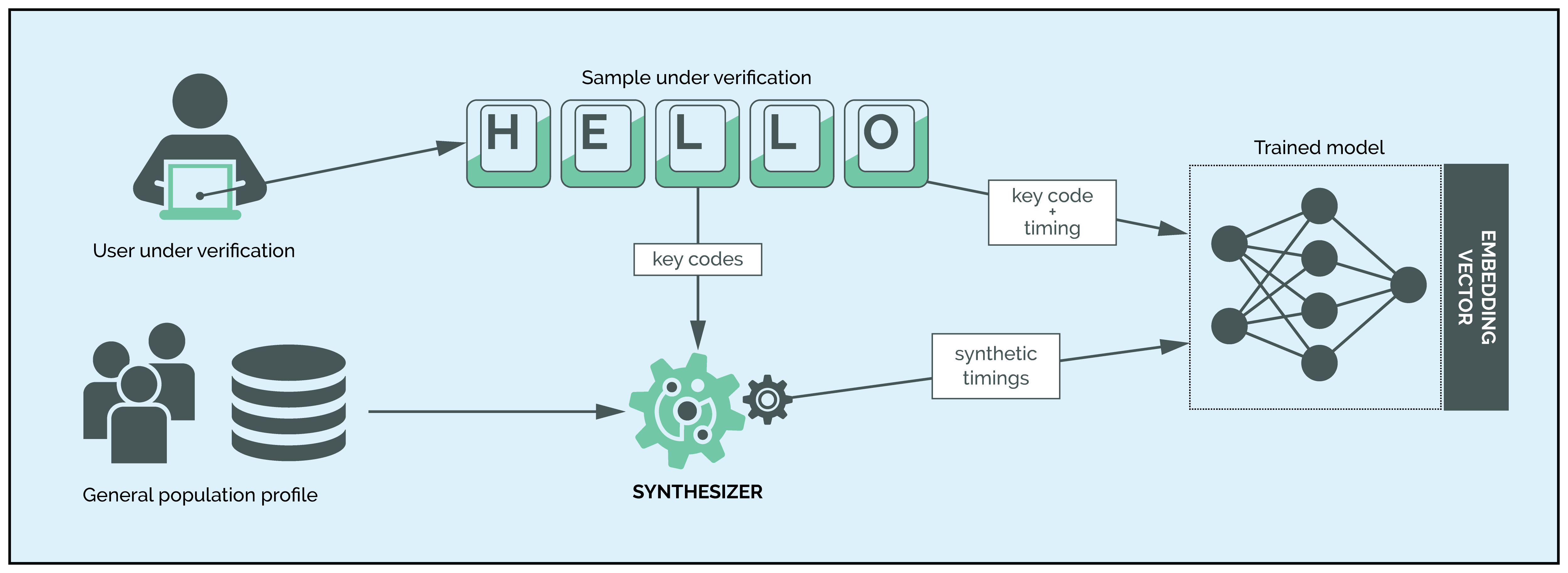}
    \caption{Representation of the feature extraction process. The key code sequence is used to generate synthetic timings based on the general population profile, which are considered together with the original keystroke timings.}
    \label{fig:synthesizer}
\end{figure*}

\subsection{Input Features}
\label{subsec:inputFeatures}

Fig. \ref{fig:synthesizer} shows a representation of the feature extraction process. The key code sequence of the sample is used to generate synthetic timings. For this purpose, a general population profile trained with all available samples is used. The synthetic timings, together with the original key codes and keystroke timings, are used as input features.

In total, five input features per keystroke are used to train the model: VK (i.e., the integer key code), two classical timing features, and two synthetic timing features. The classic timing features are HT, the hold time (i.e., interval between key press and release events), and FT, the flight time (i.e., interval between the previous and current key press events). The corresponding synthetic timing features, SHT and SFT, are used to reflect how the typing style of the subject differs from the general population. All timing features are scaled to seconds and clipped to a maximum value of 10. 

The two classic input features, hold time (HT) and flight time (FT), are calculated as
\begin{align}
t^{HT}_i &= t^{R}_i - t^{P}_i  \\
t^{FT}_i &= t^{P}_i - t^{P}_{i-1} \,\,\,\,\,\,\,\, \text{for $i > 0$, or $0$ otherwise}  
\end{align}
The synthetic features are calculated using the finite context modelling method \cite{gonzalez2015finite,gonzalez2022towards}, here denoted by $\mathcal{S}$. Given a target sample $\mathbf{w}$, the method $\mathcal{S}$ outputs a new sample $\mathbf{w}^S$ with the same keystroke sequence but with synthetic hold and flight times. For this purpose, it requires a profile $\mathcal{A}$, which consists of a set of samples that represent the behavior to be synthesized, i.e., a specific subject, a group of subjects, or, in the scope of this study, the general population as a whole. Symbolically,
\begin{align}
\mathbf{w}^S = \mathcal{S}(\mathcal{A},\mathbf{w})
\end{align}
where $\mathbf{w}^S$, in the same way as $\mathbf{w}$, is also a sequence $\mathbf{w^S_1}, \ldots, \mathbf{w^S_n}$ of length $n$ of tuples of the form
\begin{align}
(k_i, \, t^{SP}_i, \, t^{SR}_i)
\end{align}
Briefly, the finite context modelling method attempts to match, for each key, short keystroke sequences that precede it in the sample \textbf{w} with similar sequences that can be found in the profile $\mathcal{A}$. In this way, statistical distributions for each keystroke timing are inferred, which can then be sampled to output the final synthetic timings \cite{gonzalez2015finite,gonzalez2022towards}.

Here, the profile $\mathcal{A}$ used to generate the synthetic input features collects all the samples in the development set, with the objective of representing the average typing behavior of the entire training population. For each sample $\mathbf{w}$, the corresponding $\mathbf{w}^S$ is generated using the tool \cite{gonzalez2023ksdlsd} and the two synthetic features are calculated as
\begin{align}
s^{HT}_i &= t^{HT}_i - (t^{SR}_i - t^{SP}_i)  \\
s^{FT}_i &= t^{FT}_i - (t^{SP}_i - t^{SP}_{i-1}) 
\end{align}
The resulting feature vector for each keystroke is thus
\begin{align}
(k_i, \, t^{HT}_i, \, t^{FT}_i, \, s^{HT}_i, \, s^{FT}_i)
\end{align}
which is the input the model receives during training, validation, and evaluation. 

\subsection{Batch Structure}
\label{subsec:batchStructure}

The number $N$ of samples per subject and the number $K$ of subjects per batch is fixed, for a total of $NK$ samples per batch. During epoch zero, the $K$ subjects to be included in each batch are randomly chosen from all those available in the development set, with reposition between batches but making sure no batch includes repeated subjects. 

For epochs $m > 0$, the objective of the training curriculum is to progressively show to the model the nearest, i.e., harder to discriminate, subjects while still including enough random sets for the model to keep track of the global structure of the embedding space. For this reason, each batch includes:
\begin{itemize}
\item \emph{One sequential subject}, in the same order as they appear in the development set, without restarting the pointer to the current subject between epochs.
\item \emph{$m$ nearest subjects}, determined as those whose centers of their embeddings are the $m$ nearest to the center of the embeddings of the sequential subject. The embedding centers are calculated at the start of each epoch.
\item \emph{$K - 1 - m$ random subjects} chosen from all those available in the development set, with reposition between batches but making sure no batch includes repeated subjects.
\end{itemize}

%
%

\begin{figure*}[t]
    \centering
     \includegraphics[trim={0cm 0cm 0 0cm}, width=0.9 \linewidth]{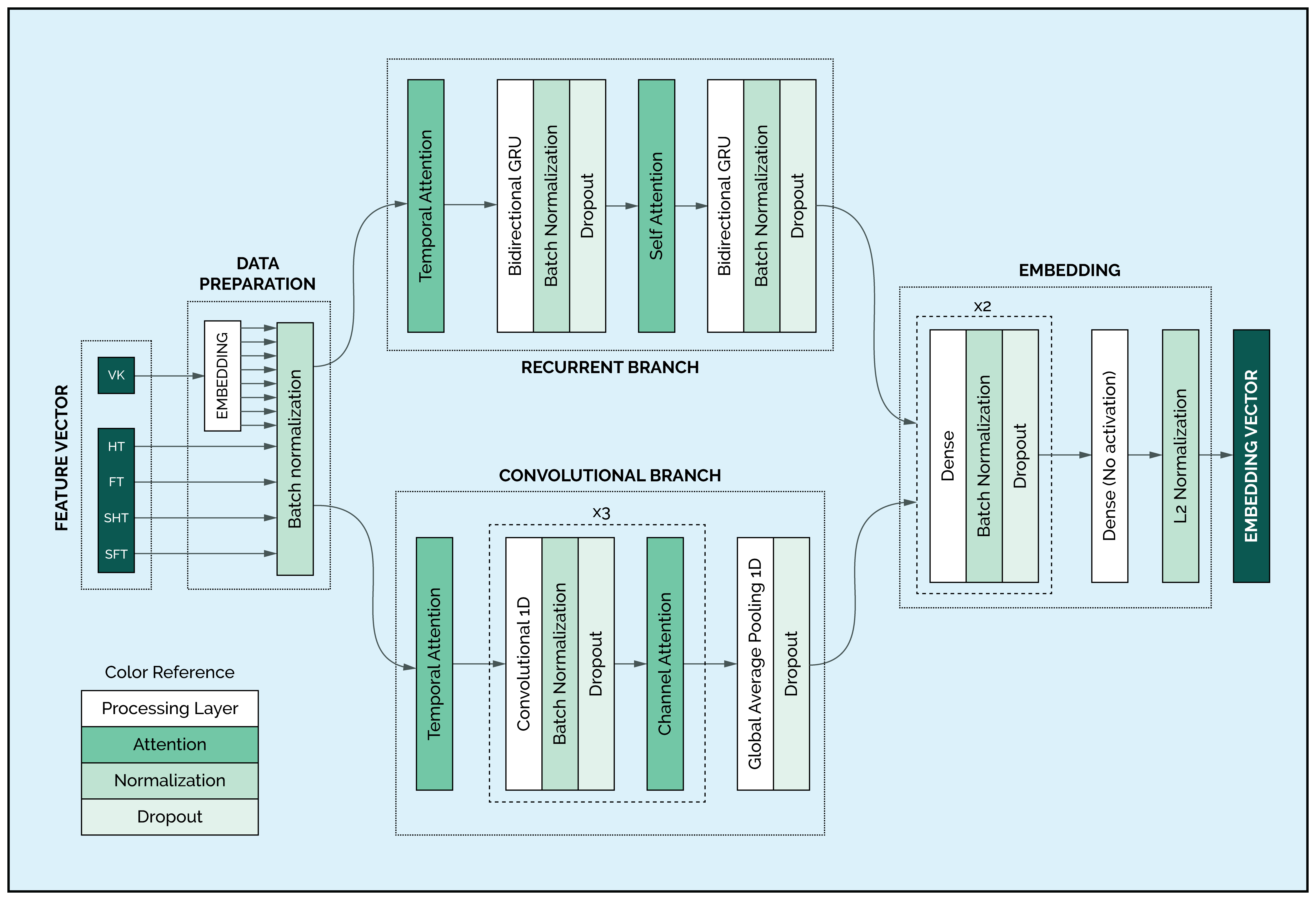}
    \caption{\catchyname: proposed dual-branch (recurrent and convolutional) embedding model for distance metric learning.}
    \label{fig:architecture}
\end{figure*}

\begin{table}[t]
    \centering
    \caption{Main hyperparameters of \catchyname{}.}
    \begin{tabular}{l|c|c}
        \hline
        \textbf{Hyperparameter}  & \textbf{Layers} & \textbf{Value}  \\
        \hline
        \hline
        Number of Units & GRU/dense & 512 \\
        Number of Units & embedding &  64 \\
        Filters & conv1D & 256, 512, 1024 \\
        Kernel Size & conv1D & 6 \\
        Activation & GRU layers & tanh \\
        Activation & conv1D/dense & ReLU \\
        Dropout Rate & & 0.5 \\
        \hline
    \end{tabular}
    \label{tab:hyperparameters}
\end{table}

\subsection{Model Architecture}
\label{subsec:modelArchitecture}

The proposed \catchyname{} architecture uses two branches, one recurrent and one convolutional, to learn how to embed the input samples into a lower dimensional space. Fig. \ref{fig:architecture} shows the details of the proposed architecture. Sample similarity is finally measured by using the Euclidean distance between their respective embeddings, as traditionally done in DML.

The choice of the architecture is motivated by the observation that keystroke timings result from a
combination of two factors: a partially conscious decision process involving \textit{what} to type and an entirely unconscious motor process pertaining to \textit{how} to type \cite{gonzalez2021shape}. The convolutional branch is expected to excel at identifying common, short sequences, while the recurrent branch is expected to capture the subjects’s time-dependent decision process. The rationale for combining both is described in Sec. \ref{subsec:ablationStudy}, where the ablation study shows that a dual-branch model outperforms a purely recurrent or convolutional single-branch model with an equivalent number of model parameters.

The proposed model is composed of four main modules: a small data preparation module whose only purpose is to embed the key code into a small dimensional space and normalize the input; the recurrent and convolutional branches that give the name to the proposed \catchyname{} model; and a final embedding module. The main hyperparameters of the model are listed in Table \ref{tab:hyperparameters}.

The recurrent branch comprises two bidirectional GRU layers (512 units), while the convolutional branch features three blocks of 1D convolution, each with an increasing number of filters (256, 512, and 1024,
with kernel size equal to 6) and utilizes global average pooling. Temporal attention serves as the first layer of both branches, i.e., scaled dot-product self attention is applied between the recurrent layers, whereas
channel attention follows each convolutional layer. 

At the embedding module, the outputs of both branches are concatenated, and the final embedding vector is produced by three dense layers. Batch normalization and dropout are
applied after each processing layer in all modules.

\begin{figure*}
    \begin{center}
\begin{tikzpicture}[scale=0.75]
    \begin{axis}[
        scatter/classes={
            U0={mark=*,draw=blue,fill=blue!50,opacity=0.6},
            U1={mark=*,draw=red,fill=red!50,opacity=0.6},
            U2={mark=*,draw=green,fill=green!50,opacity=0.6},
            U3={mark=*,draw=orange,fill=orange!50,opacity=0.6},
            U4={mark=*,draw=purple,fill=purple!50,opacity=0.6},
            U5={mark=*,draw=brown,fill=brown!50,opacity=0.6},
            U6={mark=*,draw=cyan!60!black,fill=cyan!40,opacity=0.6},
            U7={mark=*,draw=magenta,fill=magenta!50,opacity=0.6},
            U8={mark=*,draw=teal,fill=teal!50,opacity=0.6},
            U9={mark=*,draw=lime,fill=lime!50,opacity=0.6},
            U10={mark=*,draw=olive,fill=olive!50,opacity=0.6},
            U11={mark=*,draw=violet,fill=violet!50,opacity=0.6},
            U12={mark=*,draw=yellow!70!black,fill=yellow!50,opacity=0.6},
            U13={mark=*,draw=pink,fill=pink!50,opacity=0.6},
            U14={mark=*,draw=black!70!blue,fill=black!50!blue,opacity=0.6},
            U15={mark=*,draw=orange!80!black,fill=orange!50!black,opacity=0.6},
            U16={mark=*,draw=cyan!80!black,fill=cyan!50!black,opacity=0.6},
            U17={mark=*,draw=red!80!black,fill=red!50!black,opacity=0.6},
            U18={mark=*,draw=green!70!black,fill=green!50!black,opacity=0.6},
            U19={mark=*,draw=purple!70!black,fill=purple!50!black,opacity=0.6}
        },
        xmajorgrids,
        ymajorgrids,
        xticklabels=\empty,
        yticklabels=\empty,
        ylabel={t-SNE (Set2set, K=10)}
    ]
    
    \addplot[scatter,only marks,scatter src=explicit symbolic]
        coordinates {
(-17.365192,-4.7535567) [U0]
(-20.358377,-4.7280464) [U0]
(-18.144606,-5.4569273) [U0]
(-18.124678,-4.2416496) [U0]
(-20.158672,-4.3775115) [U0]
(-19.28038,-4.520717) [U0]
(-22.36744,-5.484771) [U0]
(-19.962042,-4.40484) [U0]
(-19.428556,-5.19079) [U0]
(-19.062523,-5.09465) [U0]
(-19.86511,-4.3834887) [U0]
(-18.849623,-5.2997913) [U0]
(-20.729834,-4.2369404) [U0]
(-15.936064,-4.855754) [U0]
(-20.294487,-3.8848953) [U0]
(-6.163491,-3.1086752) [U1]
(-5.622144,-2.2938359) [U1]
(-6.1057873,-4.8177304) [U1]
(5.1652737,-0.2375502) [U1]
(-6.8420486,-3.910151) [U1]
(-6.969572,-3.9227731) [U1]
(-6.9767594,-3.547274) [U1]
(-7.075829,-2.6840262) [U1]
(-7.6716213,-3.0607991) [U1]
(-6.2809424,-3.5373135) [U1]
(-7.218448,-3.7284503) [U1]
(-6.8765287,-4.6426616) [U1]
(-6.467923,-3.0835798) [U1]
(-6.977245,-4.791864) [U1]
(-6.778814,-3.1605618) [U1]
(-14.26493,2.59382) [U2]
(-14.001142,1.5580966) [U2]
(-14.517287,3.2492795) [U2]
(-13.768974,3.232042) [U2]
(-13.698449,1.5205411) [U2]
(-13.54088,2.7068348) [U2]
(-12.99132,3.0838223) [U2]
(-14.354369,3.0119014) [U2]
(-13.09217,2.7459846) [U2]
(-13.228159,2.9329731) [U2]
(-13.353332,2.3185031) [U2]
(-13.51096,2.3873348) [U2]
(-13.345266,3.4383426) [U2]
(-13.627892,2.9853516) [U2]
(-14.03261,2.7866807) [U2]
(17.409063,6.330053) [U3]
(19.505081,6.238098) [U3]
(17.47665,5.8607154) [U3]
(17.690908,5.5803103) [U3]
(18.292004,5.267656) [U3]
(17.716372,5.4511123) [U3]
(18.39367,5.8525586) [U3]
(17.85495,6.027999) [U3]
(17.290527,5.695758) [U3]
(18.535189,5.553147) [U3]
(18.178404,5.693831) [U3]
(18.48829,6.2945786) [U3]
(18.871569,6.5517716) [U3]
(17.876072,6.5028896) [U3]
(18.164024,6.704815) [U3]
(24.517538,-2.3119266) [U4]
(24.652546,-1.7946432) [U4]
(24.845339,-1.3287472) [U4]
(25.069489,-2.2027493) [U4]
(24.439884,-1.5263776) [U4]
(24.406391,-1.5483997) [U4]
(24.982422,-1.3651029) [U4]
(25.218838,-2.2837727) [U4]
(24.754711,-0.9031493) [U4]
(25.283476,-1.7448106) [U4]
(25.311941,-1.2452533) [U4]
(25.666105,-2.0543) [U4]
(25.462515,-1.5652202) [U4]
(25.057096,-1.2468498) [U4]
(24.720627,-2.187745) [U4]
(15.729071,1.2536458) [U5]
(16.75538,2.0404952) [U5]
(18.389566,1.8566767) [U5]
(18.291304,1.3775227) [U5]
(17.766783,2.1844177) [U5]
(17.855991,1.9907472) [U5]
(17.732573,1.5039158) [U5]
(17.873098,0.9625753) [U5]
(16.900118,1.582607) [U5]
(16.21821,1.7032071) [U5]
(16.811964,1.8878589) [U5]
(16.718634,1.0257566) [U5]
(15.827189,1.7006031) [U5]
(15.64968,1.7470659) [U5]
(16.401085,1.6466658) [U5]
(-5.1930842,0.28474605) [U6]
(-3.8569682,0.029755168) [U6]
(-4.8053603,-0.2495134) [U6]
(-5.106247,-0.18902768) [U6]
(-4.6906137,-0.7408354) [U6]
(-5.0839343,-0.27651912) [U6]
(-4.871628,0.6566203) [U6]
(-4.5834646,0.46131808) [U6]
(-3.9884849,0.49614456) [U6]
(-4.171694,1.1613635) [U6]
(-5.3098564,-0.09198387) [U6]
(-3.4673047,0.8248429) [U6]
(-3.4733686,0.28405234) [U6]
(-3.982674,1.2658892) [U6]
(-5.0112147,0.7346618) [U6]
(-21.645906,-7.5892) [U7]
(-22.451971,-8.163061) [U7]
(-20.73896,-6.8565593) [U7]
(-21.95652,-8.705652) [U7]
(-21.96327,-7.586927) [U7]
(-21.392256,-7.60592) [U7]
(-22.689592,-7.5820594) [U7]
(-21.595945,-8.073665) [U7]
(-22.206177,-7.5983505) [U7]
(-22.040203,-8.507936) [U7]
(-22.378836,-8.179793) [U7]
(-21.219791,-8.28956) [U7]
(-21.951828,-8.05712) [U7]
(-21.944244,-7.208712) [U7]
(-21.407736,-7.8350987) [U7]
(4.565422,2.0294235) [U8]
(2.8045964,2.172002) [U8]
(3.4187596,1.9646229) [U8]
(4.1457086,2.6936264) [U8]
(3.2038624,1.7227958) [U8]
(3.7894588,1.0564328) [U8]
(3.3300304,2.6527662) [U8]
(2.8076239,2.1721618) [U8]
(4.272339,2.5285223) [U8]
(3.8830287,2.2602813) [U8]
(3.9789941,2.062708) [U8]
(2.8675077,1.1711701) [U8]
(3.3242254,1.0140587) [U8]
(2.678786,0.7621243) [U8]
(2.5871134,1.428871) [U8]
(10.897729,8.177225) [U9]
(10.906231,8.180506) [U9]
(10.807385,8.193236) [U9]
(11.5065,7.092073) [U9]
(11.286795,7.4813843) [U9]
(10.46504,7.6107135) [U9]
(11.237289,7.632137) [U9]
(10.503224,7.675759) [U9]
(10.925191,7.101791) [U9]
(11.604838,7.771988) [U9]
(11.469231,8.397375) [U9]
(11.796326,8.156641) [U9]
(11.740638,8.095858) [U9]
(11.392068,8.394078) [U9]
(11.93919,7.2669296) [U9]
(20.181671,-0.879457) [U10]
(19.7925,-1.2559935) [U10]
(19.181265,-1.115388) [U10]
(19.657003,-0.25872654) [U10]
(19.763905,0.17529607) [U10]
(19.222828,-0.6962793) [U10]
(20.4498,-0.77571386) [U10]
(19.98797,-1.348213) [U10]
(19.725,-0.76957923) [U10]
(19.531832,-0.6142671) [U10]
(18.706108,-0.06476436) [U10]
(19.836432,-0.37144732) [U10]
(20.293682,-0.4409892) [U10]
(20.076952,-1.1188663) [U10]
(18.135899,0.78162295) [U10]
(-22.24569,-3.4570322) [U11]
(-22.536629,-4.9762726) [U11]
(-21.422445,-4.773137) [U11]
(-22.17311,-4.841411) [U11]
(-22.846138,-3.996842) [U11]
(-22.246117,-4.8690553) [U11]
(-21.065731,-4.0343547) [U11]
(-23.14824,-4.3016024) [U11]
(-23.287569,-4.203145) [U11]
(-18.452497,-5.7630377) [U11]
(-22.812698,-4.8464537) [U11]
(-22.598541,-3.635503) [U11]
(-23.795128,-4.8879857) [U11]
(-22.163704,-4.367555) [U11]
(-22.4574,-4.1244593) [U11]
(2.7197464,8.166542) [U12]
(4.4502254,9.009515) [U12]
(3.3552616,8.746574) [U12]
(3.749821,8.5379505) [U12]
(4.071135,8.413715) [U12]
(3.5795937,8.242732) [U12]
(3.3503752,7.9400973) [U12]
(3.766063,8.442683) [U12]
(3.6489792,8.978403) [U12]
(3.6900656,7.23467) [U12]
(3.4748406,7.5057216) [U12]
(3.5534065,7.405235) [U12]
(3.1937866,8.264259) [U12]
(3.5085404,9.017204) [U12]
(4.0097213,8.106502) [U12]
(21.183868,4.9897614) [U13]
(20.201292,5.271887) [U13]
(21.227,5.7250237) [U13]
(21.358553,6.2174106) [U13]
(21.57722,5.911195) [U13]
(20.682962,5.7149897) [U13]
(22.035099,5.0036225) [U13]
(20.570755,5.6495643) [U13]
(21.063221,5.598797) [U13]
(21.720657,5.8874288) [U13]
(21.394716,4.712743) [U13]
(21.447836,5.561796) [U13]
(21.593922,5.022837) [U13]
(20.90424,6.2443476) [U13]
(21.6287,4.327976) [U13]
(11.159114,2.586178) [U14]
(14.123257,3.0291245) [U14]
(13.296454,2.9274597) [U14]
(11.565918,3.6181219) [U14]
(11.32953,3.9893863) [U14]
(10.076682,2.0496767) [U14]
(10.423288,2.7708206) [U14]
(9.941407,3.029256) [U14]
(11.371283,2.7520847) [U14]
(10.749233,3.011502) [U14]
(12.707111,2.782523) [U14]
(12.43902,2.8688838) [U14]
(10.559038,3.5385902) [U14]
(10.125174,3.3669896) [U14]
(11.077355,3.58544) [U14]
(-27.136284,-6.458569) [U15]
(-26.861734,-7.055328) [U15]
(-26.52242,-6.4330416) [U15]
(-25.94714,-6.881759) [U15]
(-26.837194,-6.251376) [U15]
(-27.678398,-6.571487) [U15]
(-27.241114,-5.528368) [U15]
(-26.637241,-5.5799084) [U15]
(-27.243818,-6.344597) [U15]
(-27.382404,-5.9029846) [U15]
(-27.824924,-6.1318984) [U15]
(-27.040648,-4.7456355) [U15]
(-27.479185,-6.199454) [U15]
(-27.171637,-5.252156) [U15]
(-27.554132,-5.476359) [U15]
(-10.096675,-4.15858) [U16]
(-10.193059,-4.24791) [U16]
(-9.533049,-4.330575) [U16]
(-9.366705,-3.8075557) [U16]
(-9.959,-4.619297) [U16]
(-11.020007,-4.355748) [U16]
(-9.715604,-3.5729249) [U16]
(-10.255115,-4.7321095) [U16]
(-9.605121,-2.8131359) [U16]
(-9.705253,-4.019228) [U16]
(-7.9747624,-3.759425) [U16]
(-10.139021,-3.7693558) [U16]
(-12.12261,-3.267023) [U16]
(-11.002585,-3.5822942) [U16]
(-8.67025,-3.285775) [U16]
(-13.291689,-4.225866) [U17]
(-13.544153,-4.667353) [U17]
(-12.568622,-4.1930256) [U17]
(-12.947323,-4.252979) [U17]
(-18.92557,-3.8812985) [U17]
(-14.560793,-4.787351) [U17]
(-14.981188,-3.9524875) [U17]
(-15.276706,-5.591856) [U17]
(-14.0597515,-5.116436) [U17]
(-15.446699,-4.853555) [U17]
(-12.201028,-4.5744405) [U17]
(-15.186606,-3.7677855) [U17]
(-13.354611,-4.3703637) [U17]
(-15.071029,-4.441198) [U17]
(-12.403807,-4.5943546) [U17]
(-12.937757,-7.088184) [U18]
(-12.462167,-7.0238714) [U18]
(-12.921033,-7.5419865) [U18]
(-12.553862,-7.5630546) [U18]
(-12.44532,-7.190241) [U18]
(-13.257301,-6.163043) [U18]
(-12.207693,-6.634506) [U18]
(-12.641252,-6.9433694) [U18]
(-11.692248,-7.5541873) [U18]
(-13.224351,-7.490636) [U18]
(-12.742248,-7.6284356) [U18]
(-12.887355,-6.320461) [U18]
(-11.658433,-7.3604207) [U18]
(-11.493335,-6.4210043) [U18]
(-12.702329,-6.04049) [U18]
(6.997243,0.9385343) [U19]
(7.3127027,1.2138515) [U19]
(4.796915,1.1159685) [U19]
(6.4911914,1.6380725) [U19]
(5.436119,0.7199464) [U19]
(6.5208583,0.40990755) [U19]
(6.7584677,0.77924937) [U19]
(6.1277933,1.1140116) [U19]
(10.452209,2.5556946) [U19]
(5.9226017,0.6521562) [U19]
(6.9956956,1.4118251) [U19]
(7.383356,0.8588669) [U19]
(6.734182,0.5909109) [U19]
(6.064633,0.35642642) [U19]
(7.1390815,1.6950322) [U19]
        };
    \end{axis}
\end{tikzpicture} \begin{tikzpicture}[scale=0.75]
    \begin{axis}[
        scatter/classes={
            U0={mark=*,draw=blue,fill=blue!50,opacity=0.6},
            U1={mark=*,draw=red,fill=red!50,opacity=0.6},
            U2={mark=*,draw=green,fill=green!50,opacity=0.6},
            U3={mark=*,draw=orange,fill=orange!50,opacity=0.6},
            U4={mark=*,draw=purple,fill=purple!50,opacity=0.6},
            U5={mark=*,draw=brown,fill=brown!50,opacity=0.6},
            U6={mark=*,draw=cyan!60!black,fill=cyan!40,opacity=0.6},
            U7={mark=*,draw=magenta,fill=magenta!50,opacity=0.6},
            U8={mark=*,draw=teal,fill=teal!50,opacity=0.6},
            U9={mark=*,draw=lime,fill=lime!50,opacity=0.6},
            U10={mark=*,draw=olive,fill=olive!50,opacity=0.6},
            U11={mark=*,draw=violet,fill=violet!50,opacity=0.6},
            U12={mark=*,draw=yellow!70!black,fill=yellow!50,opacity=0.6},
            U13={mark=*,draw=pink,fill=pink!50,opacity=0.6},
            U14={mark=*,draw=black!70!blue,fill=black!50!blue,opacity=0.6},
            U15={mark=*,draw=orange!80!black,fill=orange!50!black,opacity=0.6},
            U16={mark=*,draw=cyan!80!black,fill=cyan!50!black,opacity=0.6},
            U17={mark=*,draw=red!80!black,fill=red!50!black,opacity=0.6},
            U18={mark=*,draw=green!70!black,fill=green!50!black,opacity=0.6},
            U19={mark=*,draw=purple!70!black,fill=purple!50!black,opacity=0.6}
        },
        xmajorgrids,
        ymajorgrids,
        xticklabels=\empty,
        yticklabels=\empty,
        ylabel={t-SNE (SetMarginLoss, K=2)}        
    ]
    
    \addplot[scatter,only marks,scatter src=explicit symbolic]
        coordinates {
    (-17.49278,1.8187532) [U0]
    (-18.848831,1.7789378) [U0]
    (-18.646576,3.3465362) [U0]
    (-18.769724,2.8000605) [U0]
    (-20.175062,1.5657378) [U0]
    (-18.398813,2.7369208) [U0]
    (-21.67985,0.11574431) [U0]
    (-20.522821,2.2771928) [U0]
    (-19.879799,2.570417) [U0]
    (-17.905304,2.840473) [U0]
    (-19.173388,2.2185037) [U0]
    (-19.118992,3.4135835) [U0]
    (-19.89031,2.0074978) [U0]
    (-15.735813,2.814087) [U0]
    (-19.232862,1.9960213) [U0]
    (-3.4947689,1.2683638) [U1]
    (-2.2631876,0.39964765) [U1]
    (2.1714613,0.56962967) [U1]
    (2.3470263,1.0432097) [U1]
    (-4.2123427,2.0515215) [U1]
    (-4.5284185,1.7526423) [U1]
    (-4.1720023,1.038344) [U1]
    (-3.9570532,1.7779232) [U1]
    (-4.4535904,1.1821272) [U1]
    (-3.2947042,0.9325007) [U1]
    (-3.614496,1.7251853) [U1]
    (-2.8391206,2.2256584) [U1]
    (-3.6187363,1.0559316) [U1]
    (-2.2729464,2.184668) [U1]
    (-4.133658,1.502099) [U1]
    (-24.583597,-6.1169353) [U2]
    (-14.227499,1.2281642) [U2]
    (-25.277302,-6.4136186) [U2]
    (-24.928667,-6.244255) [U2]
    (-12.769787,1.4038564) [U2]
    (-24.889467,-6.404826) [U2]
    (-24.890715,-7.1439543) [U2]
    (-24.387041,-6.3857293) [U2]
    (-24.669806,-7.029774) [U2]
    (-24.88674,-6.7978826) [U2]
    (-24.287718,-6.852799) [U2]
    (-24.100702,-6.750533) [U2]
    (-25.044914,-6.9214234) [U2]
    (-24.557352,-7.18923) [U2]
    (-24.587006,-6.443351) [U2]
    (17.97506,1.0372444) [U3]
    (15.90888,0.08203734) [U3]
    (18.361174,0.0004472474) [U3]
    (18.815105,-0.07857069) [U3]
    (19.059155,-0.41004658) [U3]
    (19.07737,0.6563025) [U3]
    (18.75509,0.75305825) [U3]
    (17.512747,1.1733209) [U3]
    (18.165028,0.9266212) [U3]
    (18.00177,0.037092812) [U3]
    (18.247051,0.3557822) [U3]
    (15.701083,0.735162) [U3]
    (15.549702,0.034531333) [U3]
    (16.818436,1.134535) [U3]
    (16.045584,1.1352849) [U3]
    (23.96927,-2.6468215) [U4]
    (24.03173,-2.257685) [U4]
    (23.82656,-1.833262) [U4]
    (24.665977,-2.4930196) [U4]
    (24.587946,-1.8395506) [U4]
    (24.69397,-1.7225207) [U4]
    (23.811043,-1.8601333) [U4]
    (24.16203,-2.6203191) [U4]
    (24.517424,-1.2332108) [U4]
    (24.659494,-2.1241674) [U4]
    (24.133034,-1.5455452) [U4]
    (23.64303,-2.5405273) [U4]
    (23.443373,-2.0949564) [U4]
    (24.202507,-1.6323488) [U4]
    (24.450592,-2.6475964) [U4]
    (20.895727,1.9557209) [U5]
    (19.461588,0.34796923) [U5]
    (20.850416,2.550976) [U5]
    (20.909178,3.7090075) [U5]
    (19.10146,2.6794672) [U5]
    (19.86098,2.2370954) [U5]
    (20.416319,4.3337) [U5]
    (20.44794,3.9513187) [U5]
    (19.599321,0.63388944) [U5]
    (19.839048,1.821537) [U5]
    (20.61457,0.8225254) [U5]
    (19.800909,1.2731482) [U5]
    (20.278097,0.7078353) [U5]
    (20.84868,1.1923993) [U5]
    (20.361021,1.2513965) [U5]
    (-2.208614,-0.23036969) [U6]
    (-2.108421,-1.5703408) [U6]
    (-2.4138143,-0.82305944) [U6]
    (-3.3258834,-0.42970583) [U6]
    (1.6246835,-0.47801945) [U6]
    (-3.5135949,-0.16129094) [U6]
    (-3.7782578,-1.0312042) [U6]
    (-2.9347723,-1.0441505) [U6]
    (-3.0644722,-1.3178954) [U6]
    (-2.3164518,-1.5742888) [U6]
    (-4.0850997,-0.32668144) [U6]
    (-2.3501637,-2.0765045) [U6]
    (-1.7102077,-0.9336923) [U6]
    (-2.575514,-2.1201298) [U6]
    (-2.9077938,-1.2394181) [U6]
    (-21.844784,3.0996258) [U7]
    (-23.340122,3.2628708) [U7]
    (-20.649075,3.049618) [U7]
    (-22.853607,2.9607995) [U7]
    (-22.231562,2.795511) [U7]
    (-22.841185,3.4490578) [U7]
    (-23.174736,2.6498768) [U7]
    (-22.31183,3.333781) [U7]
    (-22.173874,2.1828396) [U7]
    (-23.045893,3.9273672) [U7]
    (-23.146023,2.293543) [U7]
    (-22.325653,4.209706) [U7]
    (-22.753721,3.250242) [U7]
    (-22.547653,2.5711505) [U7]
    (-22.28599,3.3991318) [U7]
    (4.3700376,-0.28584906) [U8]
    (4.0055146,-1.3666892) [U8]
    (3.5166175,-0.9921063) [U8]
    (5.402172,-0.37221697) [U8]
    (3.6680427,0.027898088) [U8]
    (3.459247,-0.5460357) [U8]
    (5.7792315,-2.2248101) [U8]
    (3.0464733,-0.8381536) [U8]
    (5.705562,0.61570174) [U8]
    (4.2015057,-0.32247588) [U8]
    (4.9075155,0.67421544) [U8]
    (3.036899,-1.0255694) [U8]
    (3.3062117,-1.2967079) [U8]
    (-0.4475752,-0.51741105) [U8]
    (3.9502158,-0.62871474) [U8]
    (8.35271,-7.927421) [U9]
    (8.121183,-7.4122386) [U9]
    (8.351856,-7.175577) [U9]
    (9.00055,-6.607496) [U9]
    (8.778405,-7.296654) [U9]
    (8.731907,-6.519234) [U9]
    (9.492845,-6.652934) [U9]
    (8.49059,-7.413843) [U9]
    (9.157943,-6.43918) [U9]
    (9.532135,-7.460526) [U9]
    (8.632205,-7.9221263) [U9]
    (9.590888,-7.2363315) [U9]
    (9.011095,-7.4264445) [U9]
    (9.436225,-7.592332) [U9]
    (9.77885,-7.0438237) [U9]
    (21.250193,5.9443865) [U10]
    (20.704168,5.0280933) [U10]
    (21.102793,5.334316) [U10]
    (20.730625,5.591185) [U10]
    (20.827394,4.5506487) [U10]
    (21.892279,5.718168) [U10]
    (21.829622,4.8721952) [U10]
    (21.499382,5.24402) [U10]
    (21.028357,5.560849) [U10]
    (21.700771,4.834453) [U10]
    (21.116064,4.313725) [U10]
    (22.136402,5.2165737) [U10]
    (21.619133,5.8603516) [U10]
    (21.873732,5.458564) [U10]
    (20.64879,3.3616946) [U10]
    (-18.58077,0.7146366) [U11]
    (-21.009964,1.5938458) [U11]
    (-18.826523,1.531505) [U11]
    (-20.66238,1.0576439) [U11]
    (-20.118114,-0.5205933) [U11]
    (-20.29796,0.42226383) [U11]
    (-20.988037,0.15817149) [U11]
    (-20.566511,0.024239715) [U11]
    (-20.946373,-0.3860841) [U11]
    (-19.12838,1.7610449) [U11]
    (-20.886196,1.2216815) [U11]
    (-19.590319,0.35374972) [U11]
    (-21.768986,0.013706559) [U11]
    (-19.33737,1.5668483) [U11]
    (-18.33528,1.556206) [U11]
    (6.265272,1.1376964) [U12]
    (10.848283,0.59023756) [U12]
    (6.8675675,2.6467068) [U12]
    (7.792692,2.974991) [U12]
    (7.7623386,3.218672) [U12]
    (7.5884385,3.3112147) [U12]
    (7.6956067,2.6115105) [U12]
    (6.8957477,3.2409885) [U12]
    (7.5646644,2.2215283) [U12]
    (7.3974376,2.5405169) [U12]
    (7.0808887,2.8634646) [U12]
    (6.9029074,3.375139) [U12]
    (6.549245,2.5115187) [U12]
    (6.5044417,2.0972884) [U12]
    (7.7483597,3.5480313) [U12]
    (16.609121,-1.5945886) [U13]
    (15.985397,0.8121887) [U13]
    (14.786029,-0.5131232) [U13]
    (17.05393,-0.6436476) [U13]
    (17.06398,-0.7040763) [U13]
    (14.995101,0.20483615) [U13]
    (10.843837,-0.38806266) [U13]
    (16.597034,-1.1540619) [U13]
    (16.051865,-0.96329117) [U13]
    (15.88415,-0.4486378) [U13]
    (15.489745,-1.0714083) [U13]
    (15.337272,-0.116195455) [U13]
    (15.000308,-0.7560095) [U13]
    (16.697145,-0.7563415) [U13]
    (15.061041,0.6692948) [U13]
    (8.463448,-3.4855134) [U14]
    (20.012589,3.0624053) [U14]
    (9.410253,-2.4157617) [U14]
    (8.55519,-4.126641) [U14]
    (9.19432,-4.9753127) [U14]
    (5.881578,-3.5228453) [U14]
    (7.0977693,-3.7923965) [U14]
    (6.4591374,-3.6262896) [U14]
    (7.9298825,-2.7800019) [U14]
    (6.418627,-3.8789513) [U14]
    (8.533266,-3.1172543) [U14]
    (21.151495,0.5788089) [U14]
    (7.5442405,-4.131852) [U14]
    (7.437589,-3.74964) [U14]
    (8.449163,-5.1098547) [U14]
    (-24.936031,-0.1505838) [U15]
    (-25.538683,0.011161624) [U15]
    (-24.819191,0.22119297) [U15]
    (-25.646109,0.26643312) [U15]
    (-25.341452,-0.2636678) [U15]
    (-25.74258,-0.4302894) [U15]
    (-25.144627,-1.6482708) [U15]
    (-24.87838,-1.5075393) [U15]
    (-25.791618,-0.64081717) [U15]
    (-25.19339,-1.0538104) [U15]
    (-25.612064,-1.2427354) [U15]
    (-24.886703,-1.8374066) [U15]
    (-25.297758,-0.46238917) [U15]
    (-25.089283,-1.7271886) [U15]
    (-25.736507,-1.3216299) [U15]
    (-8.064792,2.3027122) [U16]
    (-8.665868,2.0205646) [U16]
    (-6.7229733,2.4438741) [U16]
    (-5.3203974,2.3859496) [U16]
    (-8.038176,1.8688208) [U16]
    (-8.122912,2.6735554) [U16]
    (-7.075966,1.7333192) [U16]
    (-8.756653,2.6974282) [U16]
    (-7.5934157,1.6829954) [U16]
    (-6.932138,2.1627362) [U16]
    (-5.039243,1.7413927) [U16]
    (-7.352399,2.4400177) [U16]
    (-7.095579,0.96824485) [U16]
    (-7.784522,2.032153) [U16]
    (-6.0661345,1.2800634) [U16]
    (-12.060803,2.9002018) [U17]
    (-12.579737,2.8439016) [U17]
    (-13.013334,2.65167) [U17]
    (-11.830495,2.5486772) [U17]
    (-17.439363,2.3480308) [U17]
    (-14.78064,3.3176203) [U17]
    (-13.227099,2.0352397) [U17]
    (-14.526876,2.83361) [U17]
    (-15.142294,2.7031243) [U17]
    (-14.178457,3.9752994) [U17]
    (-10.90938,1.6847807) [U17]
    (-11.601718,2.0290933) [U17]
    (-11.630835,2.9420931) [U17]
    (-14.293898,2.8229942) [U17]
    (-11.036631,2.6062806) [U17]
    (-12.193789,4.434651) [U18]
    (-11.903276,3.712188) [U18]
    (-12.795406,4.3011127) [U18]
    (-12.174322,4.1281857) [U18]
    (-11.824045,4.0246663) [U18]
    (-13.320911,4.939098) [U18]
    (-11.053059,2.9682543) [U18]
    (-13.04149,4.4309745) [U18]
    (-9.645067,2.753959) [U18]
    (-11.450641,3.5353944) [U18]
    (-12.810354,3.9184337) [U18]
    (-12.482618,3.3477786) [U18]
    (-12.333781,1.6172885) [U18]
    (-11.6974,1.7194495) [U18]
    (-13.276333,3.9134612) [U18]
    (8.515749,-1.120666) [U19]
    (8.629225,-0.32240584) [U19]
    (6.4133415,0.608407) [U19]
    (7.055215,-0.56276274) [U19]
    (6.3289065,-1.2363234) [U19]
    (8.006979,-0.036542375) [U19]
    (7.9660044,-1.0941497) [U19]
    (6.8042164,-1.4875628) [U19]
    (7.0375166,-1.3008779) [U19]
    (6.565251,-0.857035) [U19]
    (7.0403194,-0.11305365) [U19]
    (8.9388,-1.3694948) [U19]
    (7.4067945,-0.41216555) [U19]
    (5.361635,-0.061199464) [U19]
    (10.236539,0.09581583) [U19]
        };
    \end{axis}
\end{tikzpicture} \begin{tikzpicture}[scale=0.75]
    \begin{axis}[
        scatter/classes={
            U0={mark=*,draw=blue,fill=blue!50,opacity=0.6},
            U1={mark=*,draw=red,fill=red!50,opacity=0.6},
            U2={mark=*,draw=green,fill=green!50,opacity=0.6},
            U3={mark=*,draw=orange,fill=orange!50,opacity=0.6},
            U4={mark=*,draw=purple,fill=purple!50,opacity=0.6},
            U5={mark=*,draw=brown,fill=brown!50,opacity=0.6},
            U6={mark=*,draw=cyan!60!black,fill=cyan!40,opacity=0.6},
            U7={mark=*,draw=magenta,fill=magenta!50,opacity=0.6},
            U8={mark=*,draw=teal,fill=teal!50,opacity=0.6},
            U9={mark=*,draw=lime,fill=lime!50,opacity=0.6},
            U10={mark=*,draw=olive,fill=olive!50,opacity=0.6},
            U11={mark=*,draw=violet,fill=violet!50,opacity=0.6},
            U12={mark=*,draw=yellow!70!black,fill=yellow!50,opacity=0.6},
            U13={mark=*,draw=pink,fill=pink!50,opacity=0.6},
            U14={mark=*,draw=black!70!blue,fill=black!50!blue,opacity=0.6},
            U15={mark=*,draw=orange!80!black,fill=orange!50!black,opacity=0.6},
            U16={mark=*,draw=cyan!80!black,fill=cyan!50!black,opacity=0.6},
            U17={mark=*,draw=red!80!black,fill=red!50!black,opacity=0.6},
            U18={mark=*,draw=green!70!black,fill=green!50!black,opacity=0.6},
            U19={mark=*,draw=purple!70!black,fill=purple!50!black,opacity=0.6}
        },
        xmajorgrids,
        ymajorgrids,
        xticklabels=\empty,
        yticklabels=\empty,
        ylabel={t-SNE (Semi-hard Triplet Loss, K=2)}
    ]
    
    \addplot[scatter,only marks,scatter src=explicit symbolic]
        coordinates {
(-16.80385,0.42600363) [U0]
(-20.640087,0.3435397) [U0]
(-16.463444,-2.201588) [U0]
(-18.899426,-2.745797) [U0]
(-22.031391,1.6146437) [U0]
(-21.713438,0.16775976) [U0]
(-21.186453,2.729632) [U0]
(-20.95394,-0.10119313) [U0]
(-21.755667,0.5545658) [U0]
(-21.917196,-0.110492714) [U0]
(-21.160606,1.4785717) [U0]
(-20.410688,-1.539143) [U0]
(-20.706278,-0.76966757) [U0]
(-13.168504,2.1355908) [U0]
(-18.975903,1.9241298) [U0]
(-2.5828357,1.4599177) [U1]
(-2.6101146,2.498563) [U1]
(-1.5333842,4.5567913) [U1]
(-0.35880086,2.7876136) [U1]
(-1.6499183,1.6059136) [U1]
(-3.1109211,2.6387904) [U1]
(-2.297967,1.0039281) [U1]
(-1.3452281,0.94458693) [U1]
(-3.934834,3.4211104) [U1]
(-2.4608016,4.1686063) [U1]
(-2.0558126,3.2208798) [U1]
(-1.0832995,3.1605575) [U1]
(-2.164982,2.68275) [U1]
(-1.0178896,2.8837779) [U1]
(-2.1174643,2.068984) [U1]
(-15.664171,8.291744) [U2]
(-16.145498,2.809126) [U2]
(-15.567667,9.29435) [U2]
(-15.812912,9.409903) [U2]
(-16.047476,2.6587267) [U2]
(-16.75615,8.335202) [U2]
(-17.098824,9.014368) [U2]
(-15.92204,8.795294) [U2]
(-16.114958,8.1307535) [U2]
(-16.34772,8.869224) [U2]
(-16.556871,9.034765) [U2]
(-17.195007,8.629556) [U2]
(-16.196402,9.527608) [U2]
(-15.498962,8.712019) [U2]
(-16.465338,8.344089) [U2]
(14.682108,1.559248) [U3]
(13.3047905,-0.20543611) [U3]
(14.507535,2.8082907) [U3]
(14.520814,2.2396579) [U3]
(16.477493,3.2628605) [U3]
(15.935679,3.7182105) [U3]
(15.25353,4.0749497) [U3]
(15.036761,4.245004) [U3]
(14.782245,3.551176) [U3]
(16.037395,3.145042) [U3]
(15.886569,3.460758) [U3]
(14.958601,1.2503333) [U3]
(13.244333,1.0241749) [U3]
(14.222043,4.4372325) [U3]
(11.425197,4.4873185) [U3]
(22.709164,5.669456) [U4]
(23.137178,6.5700846) [U4]
(23.012447,5.2705536) [U4]
(24.3374,5.4283085) [U4]
(23.498003,6.0375566) [U4]
(23.726376,4.9049845) [U4]
(22.705666,5.9632854) [U4]
(23.450771,5.1301684) [U4]
(23.878565,5.6378937) [U4]
(23.493773,6.763946) [U4]
(23.25185,5.82016) [U4]
(22.733103,6.9234676) [U4]
(22.525257,6.2623305) [U4]
(23.93026,6.1653895) [U4]
(23.94826,6.5343103) [U4]
(21.83102,2.820093) [U5]
(20.699217,3.5830095) [U5]
(18.098309,-4.475892) [U5]
(17.498615,-5.6375575) [U5]
(15.946432,-3.9442408) [U5]
(17.19838,-5.0951962) [U5]
(16.38279,-5.5004516) [U5]
(16.140024,-5.4376125) [U5]
(20.609324,3.1129982) [U5]
(22.03744,3.4700694) [U5]
(20.477617,3.1837506) [U5]
(21.518671,3.027828) [U5]
(22.326927,4.524894) [U5]
(20.971313,2.94224) [U5]
(21.118462,3.165124) [U5]
(-3.2067494,-3.2847614) [U6]
(-4.465222,-2.6663775) [U6]
(-2.2554822,-2.0181413) [U6]
(-2.5754476,-1.9813592) [U6]
(-0.9774002,-0.75576687) [U6]
(-3.4086924,-2.1678238) [U6]
(-4.1587725,-1.6784959) [U6]
(-3.732352,-1.3859738) [U6]
(-4.448957,-2.1275902) [U6]
(-6.2921004,-2.226904) [U6]
(-3.585759,-1.6022774) [U6]
(-6.010189,-2.7129545) [U6]
(-2.9105444,-2.5945685) [U6]
(-6.932597,-2.6863935) [U6]
(-3.920658,-2.4527442) [U6]
(-20.380674,-2.643057) [U7]
(-22.259014,-1.2557862) [U7]
(-21.895447,-3.0154154) [U7]
(-22.872778,-1.2280873) [U7]
(-19.860577,-2.3101416) [U7]
(-21.365458,-1.655672) [U7]
(-22.338642,-2.0634704) [U7]
(-21.329208,-2.7069514) [U7]
(-21.424393,1.7201401) [U7]
(-21.113981,-3.7837772) [U7]
(-23.304138,-0.7112518) [U7]
(-19.849304,-0.7400038) [U7]
(-20.662811,-3.4172812) [U7]
(-21.165863,-1.7083442) [U7]
(-21.412224,-2.7602172) [U7]
(4.12974,0.47269475) [U8]
(1.6566936,0.048428174) [U8]
(3.542662,0.11749071) [U8]
(4.3263354,-1.0075672) [U8]
(1.8667548,0.19394554) [U8]
(4.7103305,0.9428168) [U8]
(4.2766438,-0.9139679) [U8]
(2.3954766,0.11930706) [U8]
(1.0759797,2.3760915) [U8]
(3.8163886,0.11342746) [U8]
(4.89108,0.27624395) [U8]
(-0.42650262,-2.3725605) [U8]
(0.7115719,-0.18336605) [U8]
(-0.85212934,-2.311461) [U8]
(4.6741176,1.0121535) [U8]
(15.937487,-9.123686) [U9]
(14.452689,-8.1800585) [U9]
(13.935168,-8.512106) [U9]
(16.803673,-6.6997757) [U9]
(17.56857,-8.45853) [U9]
(16.25675,-7.456695) [U9]
(15.276713,-6.649835) [U9]
(14.693503,-8.476336) [U9]
(18.325699,-7.993221) [U9]
(17.431671,-7.336484) [U9]
(15.968293,-9.182456) [U9]
(15.433724,-6.6372337) [U9]
(16.965574,-8.653515) [U9]
(17.320612,-7.5822277) [U9]
(18.936205,-7.3770432) [U9]
(20.552055,-6.0100627) [U10]
(19.94279,-5.6747174) [U10]
(19.127651,-5.5990734) [U10]
(18.895481,-5.8927574) [U10]
(19.568565,-5.964668) [U10]
(18.588854,-6.3924627) [U10]
(20.622528,-6.952877) [U10]
(20.73784,-6.188802) [U10]
(18.678726,-5.857371) [U10]
(19.509129,-6.6601896) [U10]
(17.827469,-5.4580255) [U10]
(19.817303,-7.3404922) [U10]
(20.176407,-7.0266824) [U10]
(20.081211,-6.5510015) [U10]
(17.266132,-4.4932094) [U10]
(-17.494204,-1.0263879) [U11]
(-18.645844,-2.2990203) [U11]
(-18.058659,-1.8997859) [U11]
(-20.29607,0.6267429) [U11]
(-18.826527,-0.09800307) [U11]
(-18.870466,-1.7902803) [U11]
(-19.330538,-1.8551104) [U11]
(-19.223333,-0.26252174) [U11]
(-19.042933,-0.7207365) [U11]
(-19.796394,-0.27562338) [U11]
(-19.987455,1.9321313) [U11]
(-19.09471,0.3486369) [U11]
(-23.2549,0.7915057) [U11]
(-17.174496,-0.2813022) [U11]
(-19.630377,2.0674033) [U11]
(10.161756,5.218905) [U12]
(11.269658,3.955623) [U12]
(9.801048,5.5307183) [U12]
(11.070143,6.423898) [U12]
(10.797958,5.468579) [U12]
(11.368414,5.2568173) [U12]
(11.242348,5.421639) [U12]
(10.233731,5.9281445) [U12]
(11.9022875,5.750832) [U12]
(13.300098,7.741608) [U12]
(13.037792,7.696479) [U12]
(12.607485,7.471211) [U12]
(10.508193,6.3189864) [U12]
(9.556079,5.9018545) [U12]
(11.92415,6.9337244) [U12]
(14.026082,-1.6492921) [U13]
(15.3264885,0.85064477) [U13]
(12.375896,-0.80476344) [U13]
(15.796813,-0.17589872) [U13]
(15.836094,-0.17474338) [U13]
(12.737535,-0.05812991) [U13]
(11.493834,-1.0322425) [U13]
(14.5625925,-0.013143867) [U13]
(12.939985,-0.7644374) [U13]
(15.1224985,-0.5901329) [U13]
(13.712337,-1.9193844) [U13]
(13.442372,-0.72238487) [U13]
(12.557338,-1.1334938) [U13]
(13.906754,-0.5785098) [U13]
(11.391736,-2.014235) [U13]
(11.247885,1.6366881) [U14]
(11.693601,1.7471274) [U14]
(12.97339,2.7750695) [U14]
(16.25075,-4.3613143) [U14]
(20.161314,-4.886732) [U14]
(7.6999354,3.0483458) [U14]
(9.348879,2.3704555) [U14]
(9.255495,4.328392) [U14]
(8.498479,3.2701573) [U14]
(9.959151,1.9233898) [U14]
(11.31443,0.9823452) [U14]
(21.840818,6.915259) [U14]
(10.139491,3.2730422) [U14]
(9.334522,2.699169) [U14]
(11.269179,2.436816) [U14]
(-25.846008,-4.033701) [U15]
(-26.276619,-2.7075331) [U15]
(-26.46621,-2.1870508) [U15]
(-26.733446,-2.382224) [U15]
(-26.068888,-3.1754184) [U15]
(-26.85808,-3.377299) [U15]
(-25.306908,-3.4682665) [U15]
(-25.935503,-2.090633) [U15]
(-25.917358,-4.5380416) [U15]
(-25.890278,-3.364815) [U15]
(-26.552761,-3.1119077) [U15]
(-24.821321,-2.8305445) [U15]
(-26.427261,-3.911406) [U15]
(-25.492268,-1.991216) [U15]
(-27.21039,-3.5063658) [U15]
(-4.5742936,0.89022356) [U16]
(-5.5507402,1.2332202) [U16]
(-3.7481413,2.505134) [U16]
(-3.2124348,3.5466425) [U16]
(-8.622592,1.8099943) [U16]
(-4.682615,1.6173089) [U16]
(-4.1935863,1.2397438) [U16]
(-5.430635,2.539834) [U16]
(-4.3036075,2.836719) [U16]
(-3.816862,1.5757142) [U16]
(-2.958348,3.5280042) [U16]
(-4.568831,0.77843696) [U16]
(-8.759768,2.0709043) [U16]
(-4.2640514,1.2514166) [U16]
(-4.7117486,3.246017) [U16]
(-13.554934,-0.071403675) [U17]
(-13.117693,0.042746473) [U17]
(-10.104602,1.6454597) [U17]
(-13.304858,-0.58552545) [U17]
(-21.369307,0.91264504) [U17]
(-16.174267,-0.4072418) [U17]
(-14.656922,-0.280261) [U17]
(-11.450121,1.6269865) [U17]
(-11.669682,2.2988908) [U17]
(-15.834803,-0.17120405) [U17]
(-9.758381,1.1443326) [U17]
(-13.072872,-0.6992808) [U17]
(-10.712319,1.0098014) [U17]
(-14.805941,0.72611004) [U17]
(-13.470592,-0.6013228) [U17]
(-13.662914,-1.974807) [U18]
(-11.581817,-1.5867481) [U18]
(-11.394086,-2.350085) [U18]
(-11.344181,-2.0752237) [U18]
(-13.946019,-2.192218) [U18]
(-14.206836,-1.2254486) [U18]
(-10.511908,0.26348868) [U18]
(-11.695169,-2.4508455) [U18]
(-4.667365,-0.78733957) [U18]
(-10.497243,-2.5665832) [U18]
(-12.195131,-2.9657953) [U18]
(-11.530172,0.47921836) [U18]
(-12.96405,-2.080766) [U18]
(-10.709442,1.775154) [U18]
(-12.218528,1.2078074) [U18]
(10.207276,-2.151213) [U19]
(10.3379,-1.2312734) [U19]
(6.2507205,0.84200704) [U19]
(8.738418,-0.7200497) [U19]
(8.326116,-0.36374494) [U19]
(8.826427,-1.4657702) [U19]
(8.490845,-1.7334996) [U19]
(7.198216,-1.354615) [U19]
(7.313026,1.1978672) [U19]
(8.059597,-1.6841059) [U19]
(7.352441,-0.42899787) [U19]
(10.179157,-2.2761588) [U19]
(8.0590925,-1.2991606) [U19]
(7.6047354,-1.2916135) [U19]
(9.538581,-1.0261018) [U19]
        };
    \end{axis}
\end{tikzpicture}

\end{center}
    \caption{t-SNE projections of embeddings for 20 subjects, comparing the proposed Set2set Loss with SetMargin Loss and Triplet Loss. Both the variance of the average class radius and the overlap between classes are minimized for the Set2set Loss.}
    \label{fig:embeddingsTSNE}
\end{figure*}

\subsection{\lossyname{} Loss Function - Motivation}
\label{subsec:lossFunctionMotivation}

The objective of the proposed \lossyname{} Loss function is to minimize the EER of a KD verification system, operating under the conditions of one-shot evaluation. Moreover, it is assumed that a fixed global detection threshold is used across all subjects. There is no loss of generality in this assumption, as will be shown later.

In this scenario, the KD system makes a verification decision with a single reference sample per subject. In other words, it outputs a measure of similarity between a pair of samples: one that certainly belongs to the legitimate subject, and another from the subject under scrutiny. If the similarity value is below the global threshold, the sample is flagged as legitimate, or otherwise as an impostor. 

Our initial point is the SetMargin Loss proposed by Morales \etal{} in \cite{morales2022setmargin}, which itself extends the well-known Triplet Loss function proposed by Schroff \etal \cite{schroff2015facenet}. Triplet Loss aims to minimize the distance between samples of the same class while simultaneously enforcing a separation between samples of different classes by, at least, a given margin. SetMargin Loss aims to capture better intra-class dependencies while enlarging the inter-class differences in the feature space, particularly along their boundaries where most classification errors occur, by adding the context of the set to the learning process.

We further extend the SetMargin Loss by letting it compare, simultaneously, arbitrarily sized sets of sets instead of just pairs of sets, while also including an additive penalty term to encourage the model to embed all classes within hyperspheres with similar average radii. The extension to arbitrarily sized sets of sets, coupled with an adequate learning curriculum, allows the loss function to capture both the global and local structure of the embedding space more effectively. Incorporating an additive penalty to address variations in the average radius among different classes improves the EER when using a fixed global detection threshold, driving it closer to the average EER achieved with optimal thresholds per subject. 

Fig. \ref{fig:embeddingsTSNE} shows the t-SNE projection of the embedding space for the same 20 subjects with models trained using Set2set Loss, SetMargin Loss, and Triplet Loss, and the same hyperparameters as in the ablation study included in Sec. \ref{subsec:ablationStudy}. In this example, SetMargin Loss, consistent with its design objective, achieves an overlap between classes of 9.33\% compared to 18.66\% of Triplet Loss, although it still clusters most sets near the center of the embedding space. Set2set Loss outperforms both of them with an overlap of 6.66\%, similar to the EER achieved in Sec. \ref{subsec:ablationStudy}. Thanks to the additive penalty, Set2set Loss obtains a variance of the average class radius equal to 0.87, significantly smaller to SetMargin Loss and Triplet Loss with values of 1.3 and 1.8, respectively.


%
%

\subsection{\lossyname{} Loss Function - Formulation}
\label{subsec:lossFunctionFormulation}

Let $d(\mathbf{x}_i^m,\mathbf{x}_j^n)$ be the Euclidean distance between the embedding vectors corresponding to the $i$--th sample of the $m$--th class and the $j$--th sample of the $n$--th class, and define
\begin{align}
L(\mathbf{x}_i^m,\mathbf{x}_j^m, \mathbf{x}_k^n) = d^2(\mathbf{x}_i^m, \mathbf{x}_j^m) - d^2(\mathbf{x}_i^m, \mathbf{x}_k^n) + \alpha
\end{align}
The \emph{Triplet Loss} function \cite{schroff2015facenet} with anchor $\mathbf{x}_i^m$, positive $\mathbf{x}_j^m$, and negative $\mathbf{x}_k^n$ is then
\begin{align}
\mathcal{L}_{TL}(\mathbf{x}_i^m,\mathbf{x}_j^m, \mathbf{x}_k^n) &= \max{ \{0, L(\mathbf{x}_i^m,\mathbf{x}_j^m, \mathbf{x}_k^n) \}}
\end{align}
The roles of the classes $m$ and $n$ can be made symmetric by defining
\begin{align}
\mathcal{L}_{STL}(m,n,i,j,k) &= \mathcal{L}_{TL}(\mathbf{x}_i^m,\mathbf{x}_j^m, \mathbf{x}_k^n) + \mathcal{L}_{TL}(\mathbf{x}_i^n,\mathbf{x}_j^n, \mathbf{x}_k^m)
\end{align}
and in terms of $\mathcal{L}_{STL}$, the \emph{SetMargin Loss} function \cite{schroff2015facenet} for the $m$--th and $n$--th classes is given by
\begin{align}
\mathcal{L}_{SM}(m,n) = \sum_{i=1}^N \sum_{j=i+1}^N \sum_{k=1}^N \mathcal{L}_{STL}(m,n,i,j,k)
\end{align}
Generalizing the above to the case when different classes have a varying number of samples is trivial. However, the assumption that all classes have $N$ samples simplifies the exposition and the implementation of the proposed loss function, while allowing for significant optimizations in computing time when the latter is vectorized. Now, using the previous $\mathcal{L}_{SM}$ and the radius penalty from equation (\ref{eq:radiuspenalty}), we can define the proposed \lossyname{} Loss, which has the form
\begin{align}
\label{eq:proposedloss}
\mathcal{L}_{S2S} = \beta \, \mathcal{L}_{RP} + \sum_{m=1}^K \sum_{n=m + 1}^K \mathcal{L}_{SM}(m,n)
\end{align}
where $\mathcal{L}_{RP}$ is the radius penalty defined in equation (\ref{eq:radiuspenalty}). The number of sets $K$ is a parameter to be optimized. In general, increasing $K$ improves the performance until a certain limit given by the capacity of the model.

In a practical implementation, the constant $\beta$ must be small enough so it does not interfere with the $\mathcal{L}_{SM}$ terms when encouraging the model to learn the structure of the embedding space, but large enough to slowly but surely enforce the normalization of the mean radii within the classes.

\section{Experimental Protocol}
\label{sec:Experimental_Protocol}
The proposed \catchyname{} model is designed using the development sets provided in the recently launched KVC-onGoing, which is based on the Aalto desktop and mobile keystroke databases \cite{stragapede2023bigdata, Dhakal2018, palin2019people}. 

\subsection{Model Training}

The batch structure has been described in Sec. \ref{subsec:batchStructure}. Values of $N=15$ samples per subject and $K=40$ subjects per batch are used. All sets in a batch need to have the same size to efficiently vectorize the loss function. Thus, $N=15$ was chosen because no subject has fewer than 15 samples in the datasets, while some have exactly 15. The value of $K=40$ is chosen as the largest that allows the optimized implementation to fit in the GPU memory. The pilot experiments (see Sect. \ref{subsec:ablationStudy}) show that increasing $K$ improves the classification accuracy of the trained model.

Each epoch consists of 20,000 steps for the desktop scenario and 7,000 steps for the mobile scenario. Validation loss is calculated at the end of each epoch. An early stopping strategy leveraging it, with a minimum delta of $10^{-4}$ and a patience of 12 epochs, is used to determine the optimal duration of the training, which lasted 18 epochs. Only the best model, as measured by the validation loss, is saved. 

The standard Adam optimizer is used for training the model. The learning rate is set to $10^{-4}$, while the rest of the parameters are left at default values, with $\beta_1=0.9$, $\beta_2=0.999$, and $\epsilon=10^{-7}$. No scheduling is used; the learning rate is kept fixed throughout the entire training process.

This value of $\beta$ in equation (\ref{eq:proposedloss}) is set to $0.05$ in the implementation of the loss function. As shown in Fig. \ref{fig:experimentBETA} of Sect. \ref{subsec:ablationStudy}, this value offers some improvement over 
smaller ones, but performance worsens rapidly if further increased.

\subsection{Tools and Frameworks}

\catchyname{} is trained with Keras 2.11.0 and Tensorflow 2.12, running in Python 3.10.10 and using an NVIDIA A100 40GB GPU. The \lossyname{} Loss is implemented as a TensorFlow function and optimized for computing speed, given that a naïve implementation of the deeply nested loop implicit in equation (\ref{eq:proposedloss}) is prohibitively slow even for small $K$. 

The synthetic timing features meant to reflect how the typing style of a subject differs from that of the general population were synthesized with the tool \cite{gonzalez2023ksdlsd}, which binary and source code are publicly available.

\subsection{Description of the Raw Data}
\label{subsec:Description_of_Databases}

The raw data within the Aalto keystroke databases consists of Unix timestamps capturing key press and release actions, each timestamp having a 1 ms-resolution and being associated with the specific ASCII code of the key pressed. The data collection took place through a web application in a completely unsupervised manner. Participants were instructed to read, memorize, and type English sentences of at least 70 characters presented to them on their smartphones. The volunteers were scattered in 163 countries, with English native speakers constituting around 68\% of the participant pool. 


Both databases are available for download from the CodaLab\footnote{\href{https://codalab.lisn.upsaclay.fr/competitions/14063}{https://codalab.lisn.upsaclay.fr/competitions/14063}} page of the KVC-onGoing. Within the competition, they are organized into four datasets, with some subjects being excluded due to insufficient acquisition sessions per subject (i.e., fewer than 15): desktop development dataset (115,120 subjects), mobile development dataset (40,639 subjects), desktop evaluation dataset (15,000 subjects), mobile evaluation datasets (5,000) subjects.
The subjects are assigned to each set following an open-set learning protocol, meaning the subjects in the development and evaluation sets are distinct. While a validation set is not explicitly provided, it can be derived from the development set using various training approaches. More information about the datasets can be found in \cite{stragapede2023ieee}.
\subsection{Evaluation Description}
\label{subsec:evaluation_description}


The comparison list provided in the KVC-onGoing competition specifies the comparisons to perform between samples\footnote{To verify that the score fluctuations due to the random allocation of the subjects in each dataset are not significant, in the design phase of the KVC-onGoing the choice was repeated 10 times with 10 different seeds, reporting a standard deviation of the EER typically being two orders of magnitude smaller than its mean value over the different choices \cite{stragapede2023ieee}.}. 

The total count of 1 vs. 1 sample-level comparisons is as follows:
\begin{itemize}
    \item Task 1 (Desktop): 2,250,000 comparisons, involving 15,000 subjects not present in the development set.
    \item Task 2 (Mobile): 750,000 comparisons, involving 5,000 subjects not present in the development set.
\end{itemize}

For each subject in the evaluation sets, 5 samples are used for enrollment (reference) and 10 samples are used for verification (probe). By considering all possible genuine pairwise comparisons, we obtain 50 comparison scores. These scores are then averaged over the 5 enrollment samples, resulting in 10 genuine scores per subject. Similarly, 20 impostor scores are generated per subject. The impostor samples are divided into two groups: 10 similar impostor scores, where verification samples are randomly selected from subjects of the same demographic group (same gender and age), and 10 dissimilar impostor scores, where verification samples are all randomly selected from subjects of different gender and age intervals.

Based on the described evaluation design, two approaches are followed to evaluate the system:
\begin{itemize}
    \item Global distributions (a fixed global comparison decision threshold): this approach entails using a single  global comparison decision threshold for all subjects.
    \item Mean per-subject distributions (subject-specific comparison decision thresholds): according to this approach, a different threshold is computed for each subject, considering the 30 verification scores (10 genuine scores, 20 impostor scores) as described above. 
    Then, the displayed results that are computed considering per-subject distributions are obtained as the mean of all subject-specific results. 
    This approach offers more flexibility, allowing the system to adapt to subject-specific distributions, thus typically improving the biometric verification performance.
\end{itemize}


\section{Experimental Results}
\label{sec:Experimental_Results}
\begin{table*}[ht!]
\centering
\caption{Comparison of the proposed \catchyname{} with the state of the art on the evaluation set of KVC-ongoing \cite{stragapede2023bigdata}.}
\begin{tabular}{P{0.11\textwidth}|P{0.1\textwidth}|P{0.1\textwidth}|P{0.1\textwidth}|P{0.1\textwidth}|P{0.1\textwidth}|P{0.1\textwidth}}
\multicolumn{7}{c}{Global Distributions} \\
\hline
\textbf{Experiment} & \textbf{EER (\%)} & \makecell[c]{\textbf{FNMR@0.1\%}\\ \textbf{FMR (\%)}} & \makecell[c]{\textbf{FNMR@1\%}\\ \textbf{FMR (\%)}} & \makecell[c]{\textbf{FNMR@10\%}\\ \textbf{FMR (\%)}} & \textbf{AUC (\%)} & \textbf{Accuracy (\%)} \\
\hline
\multicolumn{7}{c}{\textbf{Desktop}} \\
\hline
K. W. \cite{stragapede2023bigdata} & 5.22 & 67.86 & 27.98 & 1.62 & 98.79 & 94.78 \\
VeriKVC \cite{stragapede2023bigdata} & 4.03 & 59.05 & 18.79 & 1.05 & 99.07 & 95.97 \\
\textbf{\catchyname} & \textbf{3.33} & \textbf{44.17} & \textbf{11.96} & \textbf{0.51} & \textbf{99.48} & \textbf{96.68} \\
\hline
\multicolumn{7}{c}{\textbf{Mobile}} \\
\hline
K. W. \cite{stragapede2023bigdata} & 5.83 & 84.14 & 41.58 & 1.93 & 98.34 & 94.17 \\
VeriKVC \cite{stragapede2023bigdata} & 3.78 & 65.88 & 18.39 & 0.95 & 99.04 & 96.22 \\
\textbf{\catchyname} & \textbf{3.61} & \textbf{63.62} & \textbf{17.44} & \textbf{0.60} & \textbf{99.28} & \textbf{96.39} \\
\hline
\end{tabular}
\newline
\begin{tabular}{P{0.11\textwidth}|P{0.1\textwidth}|P{0.1\textwidth}|P{0.1\textwidth}|P{0.1\textwidth}}\multicolumn{5}{c}{Mean Per-subject 
Distributions} \\
\hline
\textbf{Experiment} & \makecell[c]{\textbf{EER (\%)}} & \makecell[c]{\textbf{AUC} \textbf{(\%)}} & \makecell[c]{\textbf{Accuracy} \textbf{(\%)}} & \makecell[c]{\textbf{Rank-1} \textbf{(\%)}} \\
\hline
\multicolumn{5}{c}{\textbf{Desktop}} \\
\hline
K. W. \cite{stragapede2023bigdata} & 1.78 & 99.59 & 95.90 & 94.04 \\
VeriKVC \cite{stragapede2023bigdata} & 1.32 & 99.71 & 96.14 & 95.67 \\
\textbf{\catchyname} & \textbf{0.77} & \textbf{99.87} & \textbf{96.43} & \textbf{98.04} \\
\hline
\multicolumn{5}{c}{\textbf{Mobile}} \\
\hline
K. W. \cite{stragapede2023bigdata} & 2.66 & 99.15 & 95.23 & 88.05 \\
VeriKVC \cite{stragapede2023bigdata} & 1.35 & 99.64 & 96.09 & 94.64 \\
\textbf{\catchyname} & \textbf{1.03} & \textbf{99.76} & \textbf{96.24} & \textbf{96.11} \\
\hline
\end{tabular}
\label{tab:vp}
\end{table*}

\subsection{Evaluation on the KVC-onGoing}

The verification performance of the proposed \catchyname{} is reported in Table \ref{tab:vp}. For all metrics, the reported values are computed on the evaluation set\footnote{A definition of the metrics adopted can be found in \cite{stragapede2023ieee}.}. Currently, the proposed \catchyname{} achieves the highest verification performance among all the KD systems proposed in the ongoing competition \cite{stragapede2023bigdata}. To provide a comparison with state-of-the-art KD systems, Table \ref{tab:vp} also features the results achieved by the VeriKVC and Keystroke Wizards (K. W. for short) teams, which respectively designed the second and third best performing systems.

As can be seen, \catchyname{} improves previous verification records in all cases, often by a significant margin. Considering global distributions, the EER obtained by \catchyname{} (3.33\%) is quite lower than the one achieved by VeriKVC (4.03\%) in the desktop case, whereas in the mobile case the difference is smaller (3.61\% vs 3.78\%). Moving along the columns of Table \ref{tab:vp}, the gap widens in the case of different operational points, e.g., False Non-Match Rate (FNMR) at 1\%, 10\% of False Match Rate (FMR), especially in the desktop case. The described trends are also consistent when analyzing the mean per-subject distributions. For instance, as shown in the bottom half of Table \ref{tab:vp}, \catchyname{} achieves a 0.77\% \textit{vs} 1.32\% EER (VeriKVC) for desktop, and 1.03\% \textit{vs} 1.35\% EER (VeriKVC) for mobile. Moreover, in contrast with all metrics which are related to the task of verification, the rank-1 metric reported is related to the task of identification, and it represents the percentage of times in which from a subject-specific pool of 21 samples (20 impostor samples and 1 genuine sample), the genuine one achieves the highest score \cite{stragapede2023ieee}. Consequently, given this formulation, it can only be applied to subject-specific distributions. Once again, \catchyname{} significantly outperforms previous approaches in the public KVC-onGoing competition. 

\subsection{Cross-database evaluation}

%
%

\begin{table}[]
    \caption{Cross-database evaluation results.}
    \centering
    \begin{tabular}{|l|c|c|}
        \cline{2-3}
        \multicolumn{1}{c}{} & \multicolumn{2}{|c|}{\textbf{Mean per-subject EER (G = 10) [\%]}} \\
        \hline
        \textbf{Evaluation Dataset}  & \makecell{\textbf{Type2Branch} \\ (desktop)} & \makecell{\textbf{Type2Branch} \\  (mobile)} \\
        \hline
        \hline
        Clarkson II \cite{clarksonII} & 9.10 & 12.27 \\
        LSIA \cite{gonzalez2023datasetLSIA} & 3.92 & 11.21 \\
        KM \cite{killourhy2012free} & 2.25 &  6.00 \\        
        PROSODY/GAY \cite{banerjee2014keystroke} & 3.25 &  10.00 \\        
        PROSODY/GUN \cite{banerjee2014keystroke} & 3.875 & 9.625 \\        
        PROSODY/REVIEW \cite{banerjee2014keystroke} & 1.625 & 9.375 \\        
        \hline
    \end{tabular}    
    \label{tab:crossDatabaseEvaluation}
\end{table}

In addition to the KVC-onGoing desktop and mobile evaluation sets, the proposed \catchyname{} is also evaluated using other traditional KD datasets.

Table \ref{tab:crossDatabaseEvaluation} shows the results of the cross-database evaluation on the Clarkson \cite{clarksonII}, LSIA \cite{gonzalez2023datasetLSIA}, KM \cite{killourhy2012free}, and PROSODY \cite{banerjee2014keystroke} databases. For this experiment, the \catchyname{} models were fine-tuned to account for varying capturing conditions and differences in key code representations across databases. Due to the small number of subjects included in these databases, a set of 10 subjects is left out from the fine-tuning, and used for evaluation. $G = 10$ reference sessions are used for enrollment. As expected, the mobile version of \catchyname{} performs worse on the desktop database. Nevertheless, the desktop model of \catchyname{}, despite slight degradation, still performs consistently well across all databases.

\subsection{Ablation Study}
\label{subsec:ablationStudy}

In this section we illustrate several experiments carried out during the model development phase to assess the impact of the different modules of \catchyname{} on the biometric recognition performance.

As the original training time of the model is prohibitive for a systematic ablation study, this one is carried out on a model with identical architecture and a similar training procedure but with a reduced number of parameters (the width of all layers was reduced from 512 units to 256), a reduced number of sets per batch ($K = 10$), and a smaller dataset comprising 1K subjects for training and 1K subjects for evaluation ($G = 1$). For the ablation study, we remove each architectural component, hierarchically: first, removing one branch at a time, and then each sub-components of each branch. The results are shown in Table \ref{tab:ablationStudy}, where the first column indicates the EER achieved by the ablated model, and the second column the relative worsening of the EER compared to the baseline. The EER is computed with mean per-subject distributions.

The ablation study confirms that a dual-branch model outperforms single branches, with the recurrent branch performing better on its own than the convolutional one. While each architectural component contributes to the performance, the most significant improvements are achieved by making the recurrent layers bidirectional. This can be explained by keystroke timings resulting from a combination of two factors \cite{gonzalez2021shape}: a motor process where previous keystrokes affect subsequent ones, captured by the forward-oriented layer, and a decision process where upcoming keystrokes influence the current ones, captured by the backwards-oriented layer.

%
%

\begin{table}[]
    \caption{Ablation study results. The dual-branch model outperforms single branches, with bidirectional recurrent layers having the most impact in the EER.}
    \centering
    \begin{tabular}{|l|c|c|}
        \hline
        \textbf{Model}  & \textbf{EER [\%]} & \textbf{Variation} \\   
        \hline
        \hline
        \textbf{Dual-Branch (Baseline)} & \textbf{6.51} & --- \\
        \hline
        No Keycode Embedding & 6.57 & +0.9\% \\
        No DATA PREPARATION Module & 6.66 & +2.3\%\\
        Single-layer EMBEDDING Module  & 7.26 & +11.5\% \\
        \hline
        \hline
        \textbf{Single Branch, Recurrent Only} & \textbf{6.88} & +5.7\% \\ 
        \hline
        Single Recurrent Layer & 7.11 & +3.3\% \\
        No Temporal Attention & 7.15 & +3.9\% \\
        No Self Attention & 7.62 & +10.8\% \\
        Unidirectional Recurrent Layers & 8.61 & +25.1\% \\
        \hline
        \hline
        \textbf{Single Branch, Convolutional Only} & \textbf{7.83} & +20.4\% \\ 
        \hline
        No Channel Attention & 7.91 &  +1.0\% \\
        No Temporal Attention & 8.20 & +4.7\% \\
        Double Convolutional Layer & 8.44 & +7.8\% \\
        Single Convolutional Layer & 13.01 & +66.16\% \\
        \hline
    \end{tabular}
    \label{tab:ablationStudy}
\end{table}

%
%

\begin{table}[b]
    \caption{Impact of novel training components.}
    \centering
    \begin{tabular}{|l|c|c|}
        \hline
        \textbf{Training Component}  & \textbf{EER [\%]} & \textbf{Variation} \\   
        \hline
        \hline
        \textbf{All Included (Baseline)} & 6.51 & --- \\
        \hline
        No Synthetic Features & 6.88 & +5.6\% \\
        No Improved Curriculum & 7.01 & +7.7\% \\
        \hline
        \hline
        \textbf{Set2Set (K=10) Loss Function} & \textbf{6.51} & --- \\
        \hline
        Triplet Loss (K=10) & 8.47 & +30.1\% \\        
        SetMargin Loss (K=2) & 11.64 & +78.80\% \\ 
        Triplet Loss (K=2) & 15.10 & +131.95\% \\         
        \hline
    \end{tabular}
    \label{tab:trainingImpact}
\end{table}

\subsection{Impact of Other Training Components}
\label{subsec:otherTrainingComponents}

This section provides some insights on the novel aspects introduced in the training of \catchyname{}. Table \ref{tab:trainingImpact} 
shows the effects of removing synthetic features and the improved curriculum, as well as the impact on the EER of varying the loss function. The synthetic features and improved training curriculum contribute modestly but positively, while the specialized loss function has the largest impact of all the analyzed components, yielding more than a 30\% improvement over Triplet Loss. Note that, by definition, SetMargin Loss requires $K=2$, while Triplet Loss has no such restriction. For a fair comparison, we include rows for Triplet Loss with both $K=2$ and $K=10$ in the table.

\begin{figure}[t!]
    \centering

    \begin{tikzpicture}[scale=0.9]
\begin{axis}[
    xlabel={$K$ (Number of Sets)},
    ylabel={Reduced Model EER [\%]},
    ylabel style={color=black},
    grid=major,
    xmin=0, xmax=40,
    ymin=6, ymax=12, 
    axis y line*=left,
]

\addplot[
    color={rgb,255:red,31;green,119;blue,180},
    mark=*,
    ] table {
    x   y
2 11.6446547955451842
3 9.23376475733112112
4 8.5339440357133749
5 7.8896527023598514
6 7.6542476253262706
7 7.21151819540109236
8 6.8275201742195298
9 6.942390517902499
10 6.51322369689635786
11 6.8961231353604294
12 6.559798036450901
13 6.5052472260374093
14 6.2641971312723763
15 6.7422571993726483
16 6.4556558208131378
17 6.7547322535029712
18 6.8912649545057058
19 6.6988135674630395
20 6.7303896666139691
21 6.8288644729847526
22 7.01328665291487352
23 6.4216284989108796
24 6.49977236689531235
25 6.9875313162823643
26 6.6420126253383335
27 7.01792546418634622
28 6.43315185415424196
29 6.9966840459513845
30 6.6179113079101457
31 7.024724379140327657
32 6.2508824155506253
33 6.9463389677083309
34 7.13413757788118064
35 7.12065293862935234
36 7.09875289564472489
37 6.5966638998240688
38 6.7465759802306855
39 6.4140686512386994
40 7.05114273800308422
    };
\label{plot_one}

\end{axis}

\begin{axis}[
    axis x line=none,
    axis y line*=right,
    ylabel={Full Model EER [\%]},
    ylabel style={color=black},
    xmin=0, xmax=40,
    ymin=3.2, ymax=3.8, 
    ylabel near ticks
]

\addplot[
    color={rgb,255:red,255;green,127;blue,14},
    mark=square*,
    ] table {
    x   y
    10  3.70
    20  3.4104
    30  3.327
    40  3.2012
    };
\label{plot_two}

\addlegendimage{/pgfplots/refstyle=plot_one}
\addlegendentry{Full Model}
\addlegendentry{Reduced Model}

\end{axis}
\end{tikzpicture}

    \caption{Effect on the EER with a global threshold when increasing $K$ (number of sets) in the loss function. We compare a reduced model (left y-axis) and the full model (right y-axis). Given enough training data, a larger $K$ improves the performance by providing a more comprehensive purview of the embedding space.}
    \label{fig:experimentK}
\end{figure}

Early in the development cycle, we observe a noticeable improvement on the model's performance when the number of sets simultaneously considered by the \lossyname{} Loss function increases. As Fig. \ref{fig:experimentK} shows in blue, the reduced model experiences noticeable reductions in the global EER as the value of $K$ is increased from 2 to 10. Although the performance declines for $K > 10$ in this case,  a larger model trained with a larger dataset  benefits from increasing $K$ further, as is shown in orange for the full model. It is plausible that the performance can be improved further given additional GPU memory, which would allow even larger values of $K$. 

\begin{figure}
    \centering

    \begin{tikzpicture}[scale=0.9]

\begin{axis}[
    xlabel={$\beta$ (Loss Function Hyperparameter)},
    ylabel={Global EER [\%]},
    ylabel style={color=black},
    grid=major,
    xmin=0, xmax=0.1,
    ymin=5, ymax=8, 
    axis y line*=left,
    legend pos=south east
]

\addplot[
    color={rgb,255:red,31;green,119;blue,180},
    mark=*,
    ] table {
    x   y
    0.00	7.5488
    0.01	7.3929
    0.02	6.9212
    0.03	6.7550
    0.04	6.4995
    0.05	6.5137
    0.06	6.8282
    0.07	7.3712
    0.09    7.8151
    };
\label{betaGlobalEER}

\end{axis}

\begin{axis}[
    axis x line=none,
    axis y line*=right,
    ylabel={Mean Per-Subject EER [\%]},
    ylabel style={color=black},
    xmin=0, xmax=0.1,
    ymin=5, ymax=8, 
    ylabel near ticks
]

\addplot[
    color={rgb,255:red,255;green,127;blue,14},
    mark=square*,
    ] table {
    x   y
    0.00	6.2771
    0.01	6.0303
    0.02	5.9264
    0.03	5.4053
    0.04	5.0152
    0.05	5.0036
    0.06	5.2527
    0.07	6.4498
    0.09    7.4346
    };
\label{betaAverageEER}

\addlegendimage{/pgfplots/refstyle=betaGlobalEER}
\addlegendentry{Mean P.-S. EER}
\addlegendentry{Global EER}

\end{axis}
\end{tikzpicture}

    \caption{Effect of $\beta$ on the global (left y-axis) and mean per-subject EER (right y-axis). Increasing $\beta$ improves the EER by uniformizing class radii, until this secondary objective interferes with class separation.}
    \label{fig:experimentBETA}
\end{figure}

The effect of varying $\beta$ in equation (\ref{eq:proposedloss}) is shown in Fig. \ref{fig:experimentBETA}. Both the mean per-subject and global EER benefit from increasing $\beta$ up to a range of 0.04--0.05, but beyond that the performance worsens rapidly. This can be explained by the lack of uniformity in the natural typing variation of subjects. The secondary objective of equalizing the average radius of all subjects competes with the primary objective of separating the classes. 

Another interesting aspect to evaluate is the extent to which \catchyname{} can operate with fewer training data. The smaller version of \catchyname{} employed in these experiments has originally been trained on 1,000 subjects, obtaining an EER of 6.51\%. By doubling the number of training subjects, in Table \ref{tab:numberSubjects} we observe a significant reduction of the EER (4.56\%). After that, the EER keeps reducing, but no so significantly.

\begin{table}[]
    \caption{Impact of increasing the number of subjects used for training the model.}
    \centering
    \begin{tabular}{|c|c|}
        \hline
        \makecell{\textbf{Number of}\\ \textbf{Subjects}}  & \textbf{EER [\%]} \\         
        \hline
        1,000 & 6.51 \\
        2,000 & 4.56\\
        5,000 & 4.43 \\
        25,000 & 4.07 \\
        \hline
    \end{tabular}
    \label{tab:numberSubjects}
\end{table}

\subsection{Limitations}
\label{subsec:limitations}

The proposed \catchyname{} has been trained on a balanced subset of the Aalto datasets, which consist of typing samples from transcribed text. Unfortunately, there are  no publicly available datasets comprising free-text samples that are sufficiently large for training and evaluating a model of this size and complexity.

Furthermore, the evaluation of the proposed model was conducted using an attack model that assumes zero-effort impostors, meaning that the impostor samples do not exhibit a deliberate attempt to emulate the typing style of the legitimate users. While it is standard practice to assess behavioral biometric systems under this assumption, conducting evaluations under more advanced attack models could potentially reveal decreased performance. However, to the best of our knowledge, there is currently no publicly available database containing trained impostors that could address this gap, either for quantifying the resilience of the proposed model or for retraining it with liveness detection in mind.

\section{Conclusions and Future Work}
\label{sec:Conclusions_and_Future_Work}

In the present study, we have proposed and evaluated the performance of \catchyname, a dual-branch (recurrent and convolutional) distance metric learning model for the task of verifying subjects based on their keystroke dynamics. Our approach leverages large databases to train a deep learning model that can scale to hundreds of thousands of subjects with little performance loss.

To the best of our knowledge, the proposed \catchyname{} outperforms the state of the art with a mean per-subject EER of 0.77\% in the desktop scenario and 1.03\% in the mobile scenario when using 5 enrollment samples. Under a harder evaluation criteria, when only a single enrollment sample per subject is available and a fixed global threshold is used, the proposed model still manages to achieve an EER of 3.33\% in the desktop scenario and 3.61\% in the mobile scenario. These results have been validated with a sound experimental protocol and using publicly available training and evaluation datasets during the KVC-onGoing competition \cite{stragapede2023bigdata}. 

The ablation study has shown that a dual-branch architecture outperforms single-branch architectures for a given number of parameters. Other factors such as the proposed synthetic features and the improved training curriculum also provide cumulative gains. Nevertheless, the most noticeable improvement was provided by the \lossyname{} Loss that, by extending SetMargin Loss to larger numbers of sets, allows the model to optimize the embedding space globally and results in noticeable performance gains.


Future work will be oriented towards exploring architectures with more than two branches, improving the branches of the proposed model, and evaluating the generalizability and utility of the \lossyname{} Loss function for other biometric verification and general classification tasks. This will include evaluating its resilience against mimicry, data poisoning, and other attack vectors.

\section*{Acknowledgments}
This research has received funding from projects INTER-ACTION (PID2021-126521OB-I00 MICINN/FEDER) and PowerAI+ (SI4/PJI/2024-00062, funded by Comunidad de Madrid through the grant agreement for the promotion of research and technology transfer at UAM), and by Comunidad de Madrid (ELLIS Unit Madrid). 
The authors would like to thank Ms. Susan Essex for proofreading and language editing this manuscript, and Brian Callipari for providing the illustrations.

\newpage 

\bibliographystyle{IEEEtran}
\bibliography{0_Main,NG}

 

\newpage

\begin{IEEEbiography}[{\includegraphics[width=1in,height=1.25in,clip,keepaspectratio]{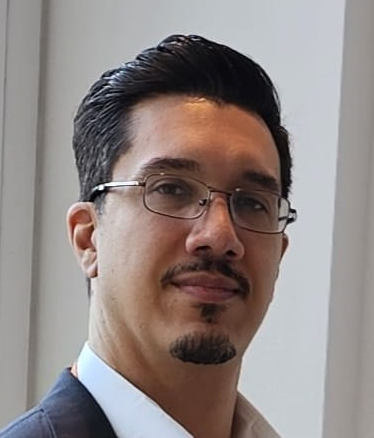}}]{Nahuel González} received his PhD degree in Computing Science from Universidad Nacional de La Plata (UNLP), Argentina, in 2022. His PhD thesis was awarded the Raúl Gallard Prize, handed by the Network of Argentinian Universities of Computing Science (RedUNCI), the following year. He has been affiliated with the Laboratorio de Sistemas de Información Avanzados (LSIA) of the University of Buenos Aires (UBA) since 2013. His main research interests are behavioral biometrics and time series prediction/classification using deep learning. He is a member of the editorial board of Data in Brief, Elsevier, since 2024.
\end{IEEEbiography}

\vspace{-5mm}

\begin{IEEEbiography}[{\includegraphics[width=1in,height=1.25in,clip,keepaspectratio]{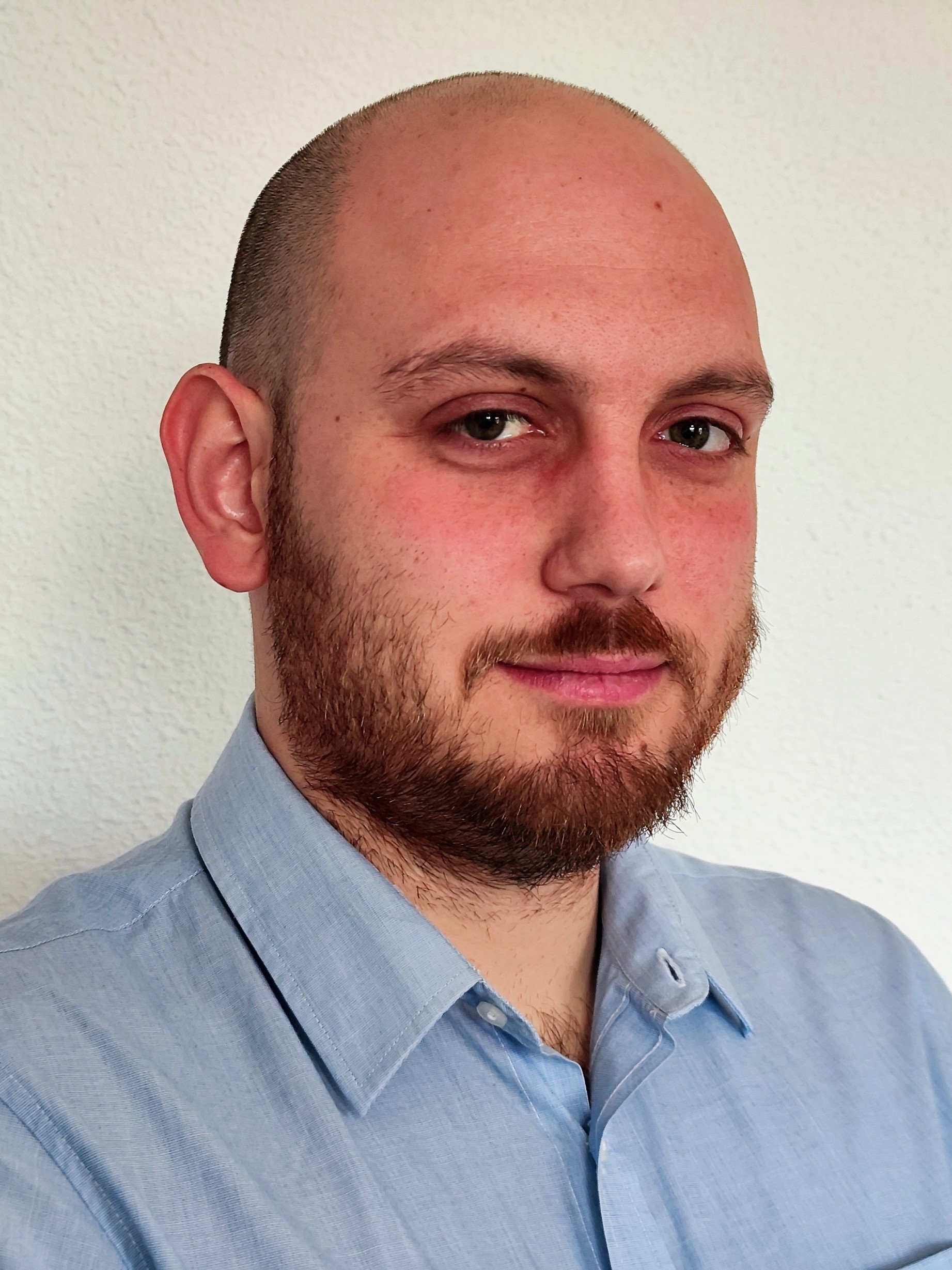}}]{Giuseppe Stragapede} received his MSc degree in electronic engineering from Politecnico di Bari, Italy, in 2019. After one year as a computer vision engineer in the industry, in 2020 he started his PhD with a Marie Curie Fellowship within the PriMa (Privacy Matters) EU project in the Biometrics and Data Pattern Analytics - BiDA Lab, at the Universidad Autonoma de Madrid, Spain. His research interests include biometrics (especially mobile biometrics), data protection, signal processing, and machine learning.
\end{IEEEbiography}

\vspace{-5mm}

\begin{IEEEbiography}[{\includegraphics[width=1in,height=1.25in,clip,keepaspectratio]{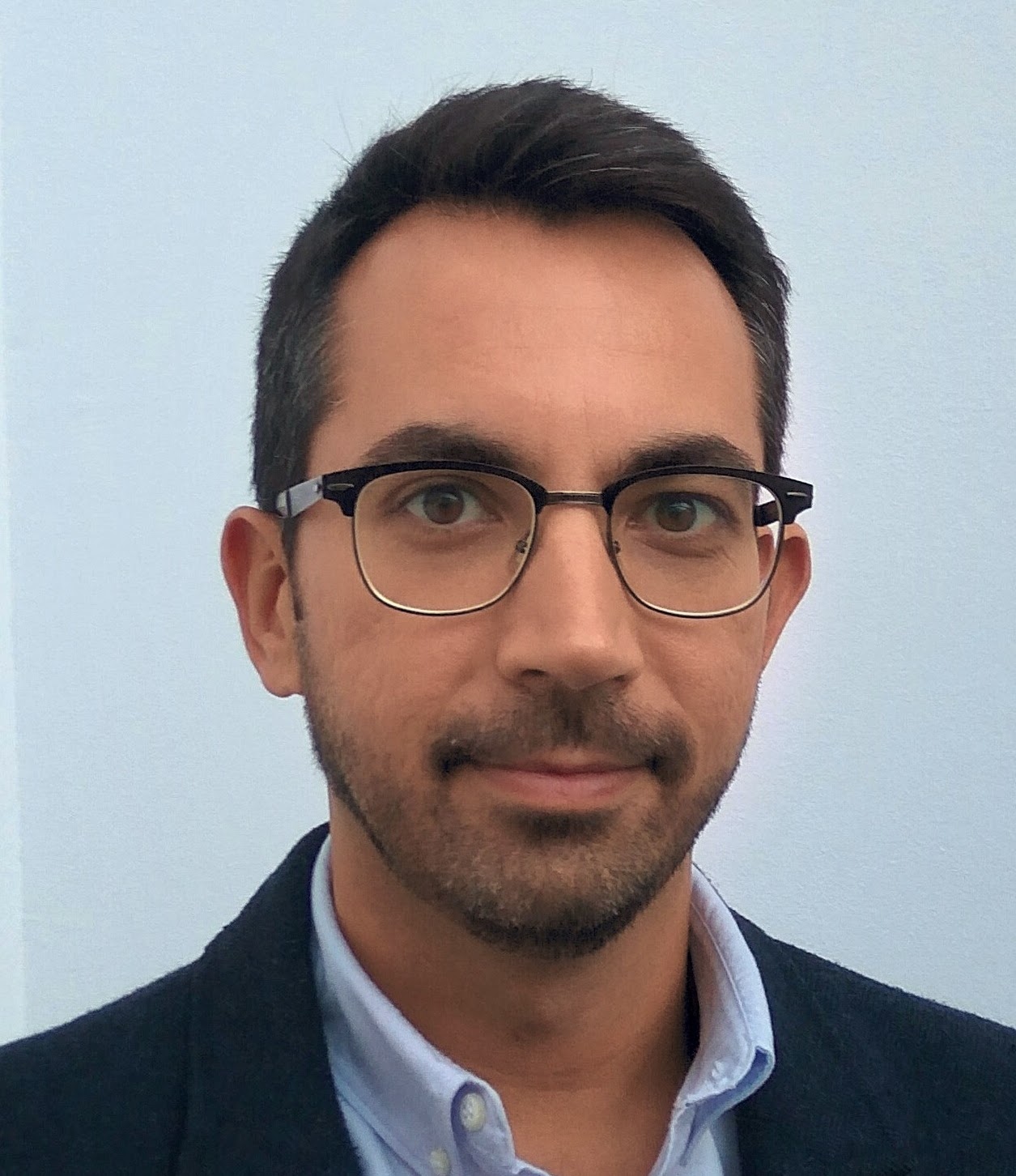}}]{Ruben Vera-Rodriguez} received his PhD degree in electrical and electronic engineering from Swansea University, U.K., in 2010. Since then, he has been affiliated with the Biometric Recognition Group, Universidad Autonoma de Madrid, Spain, where he is currently an Associate Professor since 2018. His research interests include signal and image processing, pattern recognition, HCI and biometrics, with emphasis on signature, face, gait verification, mobile biometrics and forensic applications of biometrics. He is actively involved in several national and European projects focused on biometrics. He has been awarded recently with a Medal from the Spanish Royal Academy of Engineering for his research contributions. He is member of ELLIS Society.
\end{IEEEbiography}

\vspace{-5mm}

\begin{IEEEbiography}[{\includegraphics[width=1in,height=1.25in,clip,keepaspectratio]{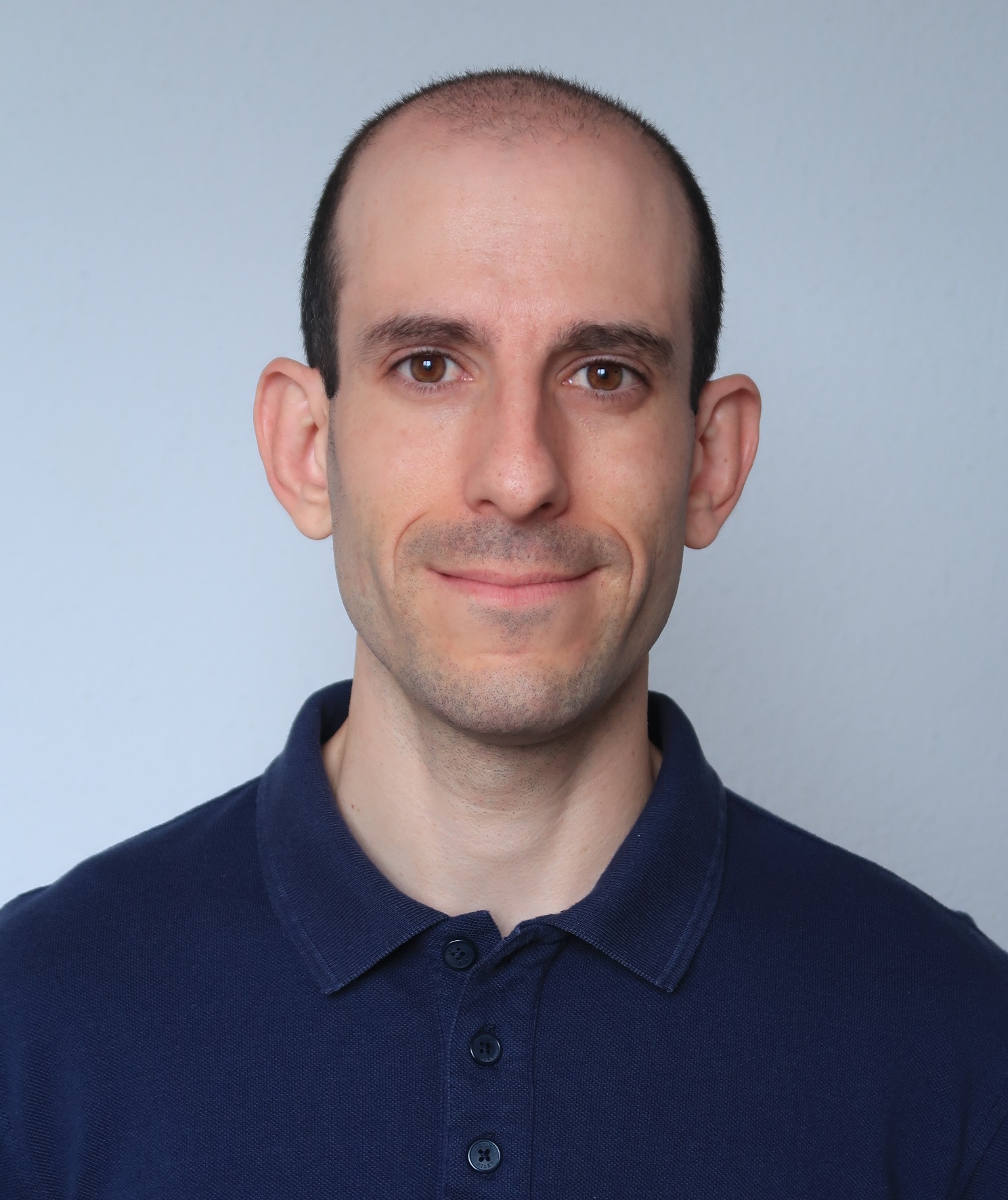}}]{Ruben Tolosana} received the M.Sc. degree
in Telecommunication Engineering, and the Ph.D. degree in Computer and Telecommunication Engineering, from Universidad Autonoma de Madrid, in 2014 and 2019, respectively. In 2014, he joined the Biometrics and Data Pattern Analytics - BiDA Lab at the Universidad Autonoma de Madrid, where he is currently an Assistant Professor. He is a member of the ELLIS Society and ELLIS Unit Madrid, Technical Area Committee of EURASIP, and Editorial Board of the IEEE Biometrics Council Newsletter. His research interests are mainly focused on signal and image processing, pattern recognition, and machine learning, particularly in the areas of DeepFakes, Human-Computer Interaction, Biometrics, and Health. He is author of more than 100 scientific articles published in international journals and conferences. He has served as General Chair and Program Chair (AVSS 2022), and Area Chair (IJCB 2023, ICPR 2022) in top conferences. Dr. Tolosana has also received several awards such as the ``European Biometrics Industry Award (2018)" from the European Association for Biometrics (EAB) and the ``Juan López de Peñalver Award" from the Spanish Royal Academy of Engineering for his research contributions and transfer of technologies.
\end{IEEEbiography}

\vfill




\end{document}